\documentclass[10pt,twocolumn,letterpaper]{article}

\usepackage[pagenumbers]{cvpr} 
\usepackage{lipsum}
\usepackage{epigraph}

\usepackage[dvipsnames]{xcolor}

\definecolor{cvprblue}{rgb}{0.21,0.49,0.74}
\usepackage{amssymb}
\usepackage[pagebackref=false,breaklinks,colorlinks,citecolor=RoyalBlue,bookmarks=false]{hyperref}

\usepackage{nicefrac}       %
\usepackage{microtype}      %

\usepackage{blindtext}
\usepackage{float}
\usepackage{enumitem}
\usepackage{tabulary,multirow,overpic}
\usepackage{verbatim}
\usepackage{color}
\usepackage{dblfloatfix}
\usepackage{pifont}%
\usepackage[accsupp]{axessibility}  %
\usepackage{makecell}

\usepackage{marvosym}

\usepackage{dblfloatfix}
\usepackage{nicefrac}

\usepackage{relsize}

\usepackage[capitalize]{cleveref}
\crefname{figure}{Figure}{Figures}

\usepackage{listings}
\definecolor{linkcolor}{RGB}{255,0,0}
\definecolor{urlcolor}{RGB}{255,105,180}
\definecolor{citecolor}{RGB}{66,168,235}
\definecolor{codegreen}{rgb}{0,0.5,0}
\definecolor{codeblue}{rgb}{0.25,0.5,0.5}
\definecolor{codegray}{rgb}{0.6,0.6,0.6}

\newlength\savewidth

\newcolumntype{x}[1]{>{\centering\arraybackslash}p{#1pt}}
\newcolumntype{y}[1]{>{\raggedright\arraybackslash}p{#1pt}}
\newcolumntype{z}[1]{>{\raggedleft\arraybackslash}p{#1pt}}

\definecolor{baselinecolor}{gray}{.92}

\definecolor{demphcolor}{gray}{.2}

\definecolor{demphcolorinline}{gray}{.3}

\definecolor{demphcolor1}{gray}{.6}

\newcommand{\ours}{{LVM}\xspace}

\title{Sequential Modeling Enables Scalable Learning for Large Vision Models}

\begin{document} 
\author{
Yutong Bai\textsuperscript{1,2,*},\enspace 
Xinyang Geng\textsuperscript{1,}\thanks{Equal Contribution, work done while at BAIR. Further updates available at \url{https://yutongbai.com/lvm.html}. },\enspace
Karttikeya Mangalam\textsuperscript{1},\enspace
Amir Bar\textsuperscript{1},\\
Alan Yuille\textsuperscript{2},\enspace
Trevor Darrell\textsuperscript{1},\enspace
Jitendra Malik\textsuperscript{1},\enspace
Alexei A. Efros\textsuperscript{1}\\
\\
\textsuperscript{1}UC Berkeley (BAIR) \qquad \qquad \textsuperscript{2}Johns Hopkins University 
}

\maketitle

\begin{abstract}
We introduce a novel sequential modeling approach which enables learning a Large Vision Model ({\ours}) without making use of any linguistic data. 
To do this, we define a common format, ``visual sentences", in which we can represent raw images and videos as well as annotated data sources such as semantic segmentations and depth reconstructions without needing any meta-knowledge beyond the pixels.  Once this wide variety of visual data (comprising $420$ billion tokens) is represented as sequences, the model can be trained to minimize a cross-entropy loss for next token prediction. By training across various scales of model architecture and data diversity, we provide empirical evidence that our models scale effectively. Many different vision tasks can be solved by designing suitable visual prompts at test time. 
\end{abstract}
\section{Introduction}

\label{sec:intro}

Large language models (LLMs) such as GPT~\cite{gpt3} and LLaMA~\cite{touvron2023llama}  have taken the world by storm. What would it take to build a Large Vision Model (LVM)?  From the animal world, we know that visual competences are not dependent on language. In particular, many experiments have shown that the visual world of non-human primates is remarkably similar to that of humans. So while the space of vision-language models such as LLaVA~\cite{llava} is interesting and worthwhile to pursue, in this paper we seek an answer to a different question – how far can we go from pixels alone?

The key features of contemporary LLMs that we seek to emulate in LVMs are: 1) scaling in the presence of big data, and 2) flexible specification of tasks through prompting (in-context learning). How do we achieve this?  As usual, there are three main components that must be specified:

{\bf Data:}  We want to exploit all the remarkable diversity in visual data. First of all, just raw unannotated images and videos. Next, we want to exploit the variety of annotated visual data sources that have been produced over the last couple of decades – semantic segmentations, depth reconstructions, keypoints, multiple views of 3D objects, among others. We define a common format, ``visual sentences'',  in which to represent these different annotations without needing any meta-knowledge beyond the pixels. The total size of our training dataset is $1.64$ billion images/frames.

{\bf Architecture:} We use a large transformer architecture ($3$ billion parameters) trained on visual data represented as sequence of tokens, using a learned tokenizer that maps each image to a string of $256$ vector-quantized tokens.

{\bf Loss function:} We draw inspiration from the natural language community, where masked token modeling has given way to sequential autoregressive prediction. Once images/videos/annotated images can all be represented as sequences, we can train the model to minimize the cross-entropy loss for predicting the next token.

With this extremely simple design, we demonstrate some noteworthy behaviors:
\begin{itemize}
\item Appropriate scaling behavior as one increases model size and data size. 
\item  Many different vision tasks can now be ``solved'' by designing suitable prompts at test time. While the results don't show as high performance as bespoke, specifically-trained models, the fact that so many tasks are all addressed by a single vision model is quite encouraging. 
\item We see a clear benefit of the amount of unsupervised data on the performance on various standard vision tasks.
\item We see a hint of an ability for general visual reasoning – handling out-of-distribution data, and performing novel tasks.  But further investigation is needed. 
\end{itemize}

\section{Related Work}
\label{sec:related}
\noindent\textbf{Pretrained Vision Models.} The value of using pretrained models (such as ImageNet-pretrained AlexNet~\cite{krizhevsky2012imagenet}) has been demonstrated as far back as 2015 in R-CNN~\cite{girshick2015region}, and it has since become standard practice in computer vision.  Self-supervised pretraining was proposed as a way to vastly increase the amount of data available for pretraining~\cite{Doersch_2015_ICCV,zhang2016colorful,noroozi2016unsupervised,pathak2016context,chen2020exploring,grill2020bootstrap}. Unfortunately, this was not very successful, likely because the CNN-based architectures of that time did not have enough capacity to absorb the data. With the introduction of Transformers~\cite{vaswani2017attention}, which have much higher capacity,  
researchers revisited self-supervised pretraining, and showed that transformer-based masked image reconstruction approaches, such as BEiT~\cite{bao2021beit}, MAE~\cite{he2021masked}, SimMIM~\cite{xie2021simmim}, perform vastly better than their CNN-based counterparts~\cite{pathak2016context}. 
Yet, despite their recent successes, current pretrained vision-only models have had trouble scaling up to the really large datasets, such as LAION~\cite{schuhmann2022laion}.

\vspace{1.3mm}
\noindent\textbf{Multi-task Learning and In-context Learning.}
From the classic one-model-per-task setups, computer vision is slowly moving toward a single model performing multiple different tasks.  
Various multi-task learning approaches~\cite{zamir2018taskonomy,kokkinos2017ubernet,doersch2017multi,sener2018multi,jaegle2021perceiver} exist but they are typically limited to a fixed, pre-defined number of tasks. 
More recently, methods inspired by in-context learning in LLMs forgo any notion of tasks and instead let the model infer the task directly from the input prompt.  For example, Visual Prompting  ~\cite{bargandelsman2022visual,wang2023images} takes in a task input/output example pair and a query image at test time, concatenates them into a single 2-by-2 image, and uses inpainting to generate the desired output. But, since the inpainting is performed using a variant of MAE~\cite{he2021masked}, the same problems with scaling are inherited by these approaches. 

\vspace{1.3mm}
\noindent\textbf{Auto-regressive Visual Models.}
The idea of using auto-regressive models for synthesizing visual data goes back at least 70 years.  Inspired by Shannon's use of $N$-grams to synthesize language~\cite{Shannon48,shannon1951prediction}, a number of works, starting with Attneave's seminal 1954 paper~\cite{attneave1954some}, applied this idea to sequentially synthesizing pixels~\cite{Garber81,Popat93,Efros99,hertzmann2001image}, image patches~\cite{Efros01}, video frames~\cite{VideoTextures}, and motion capture data~\cite{Arikan02,kovar_motion_2002,Jehee02}.  As deep models became popular, newer works replaced $N$-grams with RNNs or CNNs for pixel synthesis~\cite{pixelRNN,pixelCNN}. Most recently, transformer-based autoregessive  visual generation methods have been proposed~\citep{chen2020generative, yu2021diverse, esser2021taming, yu2021vector}, and, combined with language, have demonstrated impressive image synthesis results, e.g. Parti~\cite{parti}. 
\begin{figure*}
    \centering
    \includegraphics[width=\linewidth]{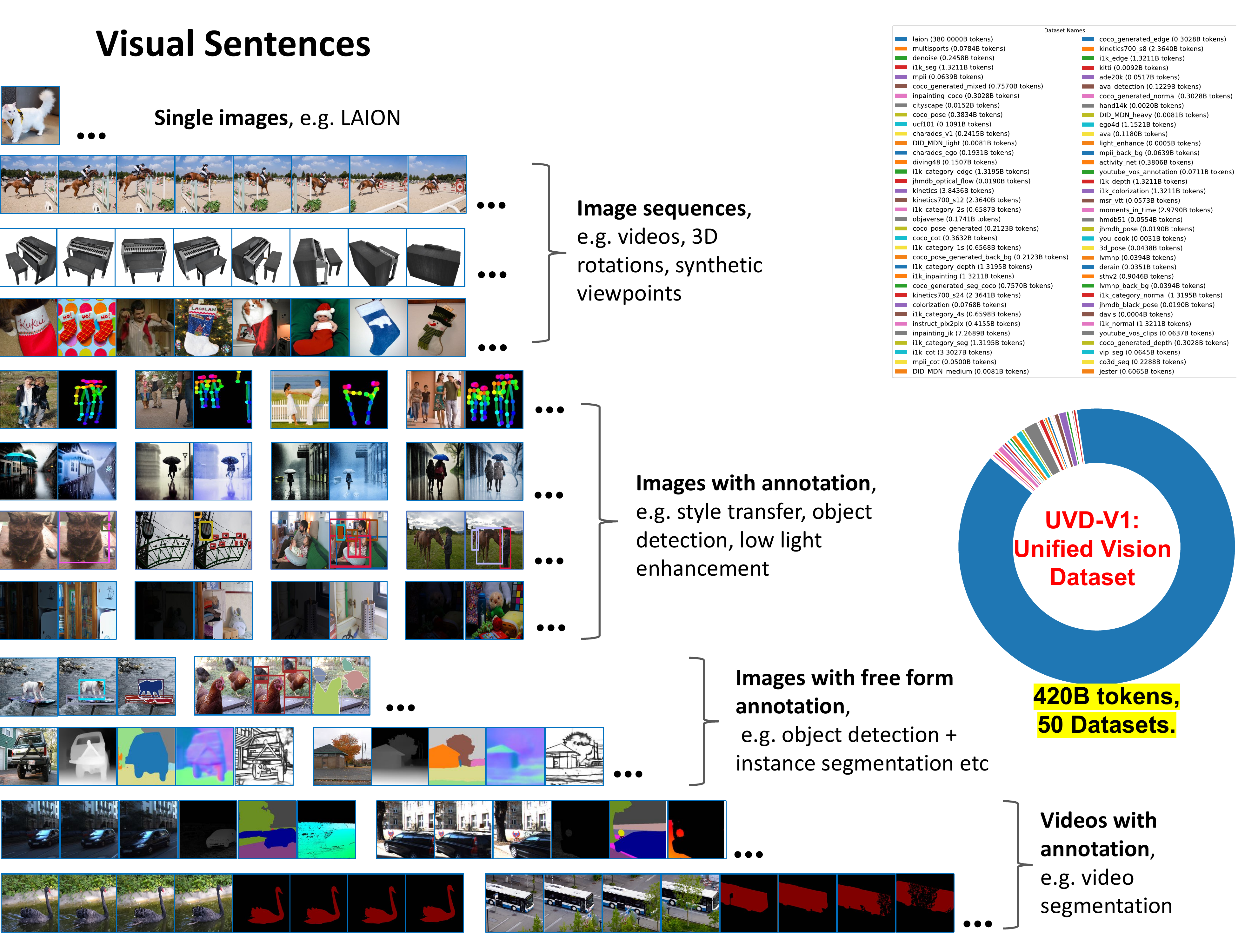}
    \vspace{-8mm}
    \caption{\textbf{Visual sentences} allow us to format diverse vision data into the unified structure of image sequences. }
    \label{fig:visual_sentences}
    \vspace{-6mm}
\end{figure*}

\section{Data}

\epigraph{\emph{``Data! Data! Data! I can't make bricks without clay!''}}{\textsc{Sherlock Holmes}}

The key requirement of any Large Pre-trained Model is that it must be trained on vast amounts of data.  For language models, very large and very diverse datasets are fairly easy to obtain.  For instance, the popular Common Crawl repository~\cite{commoncrawl} contains 250 billion web pages spanning the entire Web, is extremely diverse, and includes ``natural demonstrations'' like language translations, question answering, etc.  In computer vision, we are still very far from having a data source of comparable size and diversity.  One of the central contributions of our work is the first step toward curating such a dataset that we call {\em Unified Vision Dataset v1 (UVDv1)}.  To assemble it, we leverage many different sources of visual data: (1) unlabelled images, (2) images with visual annotations, (3) unlabelled videos, (4) videos with visual annotations, and (5) 3D synthetic objects.   
The unlabeled images, which represent over 80\% of our data, capture a huge cross-section of our visual world, and provide the required diversity, at the cost of lower quality.  Images with annotations have a much more constrained distribution, but are usually of higher quality.  Video data is even more constrained (typically, to human-centric activities), but is an invaluable source of temporal data.  Renderings of 3D synthetic objects are the lowest in diversity but can provide valuable hints about the behavior of 3D structures. 
Importantly, UVDv1 is a purely visual dataset, with no non-visual meta-data (e.g. text) included.  All together, UVDv1 contains $1.64$ billion images.   

Another important difference from Large Language Models is that language data has a natural, unified one-dimensional structure for all the data -- a stream of text.  Unfortunately, this is not the case for visual data, with different sources all having different structures.  In this work we propose {\em visual sentence} as a unified unit of visual data, which enables us to train scalable models from a diverse set sources.  A visual sentence is simply a sequence containing one or more images followed by an end-of-sentence (EOS) token. Figure~\ref{fig:visual_sentences} shows how the various data sources are partitioned into visual sentences.  In particular:

\vspace{1mm}
\noindent\textbf{Single images.} A single image itself represents the simplest form of a visual sentence -- \{ image, EOS\}. We use the filtered subset of 1.49 billion images~\cite{webster2023deduplication} from the LAION 5B~\cite{schuhmann2022laion5b} dataset.  This is by far the largest part of our data, comprising 88.5\%. 

\vspace{1mm}
\noindent\textbf{Image sequences.} A sequence of images is a natural form of visual sentence.  We create such sequences by sourcing video data from a wide range of existing datasets \cite{soomro2012ucf101, perazzi2016benchmark, kuehne2011hmdb, caba2015activitynet, monfort2019moments, monfort2021multi, reizenstein2021common, sigurdsson2016hollywood, goyal2017something, das2013thousand, carreira2019short, xu2016msr, xu2018youtube, materzynska2019jester, li2018resound, li2021multisports, sigurdsson2018charades, murray2012ava, grauman2022ego4d}. Visual sentences of 16 frames are formed by randomly sampling the videos at three different strides (10, 20, and 30).  In addition,  
we utilize synthetic 3D objects from the Objaverse Dataset~\cite{deitke2023objaverse} to generate object-centric multiview sequences for a variety of objects. 
For each object, we sample one radius length between the object center and the camera from 1.5 to 2.2, and sample one constant elevation from -45 degrees to 45 degrees, then traverse different views of the object by changing the azimuth with a step length of 15 degrees and render 24 views.  We rendered 42000 such sequences in total for training and 8000 for testing. 
Finally, we can also represent images belonging to the same semantic category as being (part of) a sequence. We use categories from ImageNet, concatenating together groups of images  (2,4,8, or 16) from the same category into a 16-image long visual sentences.  

\noindent\textbf{Images with annotations.}
To handle different types of image annotations in a uniform way, we choose to represent all annotations as images. Some data types, e.g. semantic segmentation maps~\cite{zhou2019semantic}, edge maps~\cite{xsoria2020dexined}, depth~\cite{ranftl2021vision} and normal images~\cite{bae2021estimating}, are already represented this way.  For others we apply tailored methods for each specific annotation type: 1) Object Detection: We create annotations by overlaying a color-coded bounding box around each object, following the methodology in~\cite{mmdetection}; 
2) Human Pose: Human skeletons are rendered in pixel space, adhering to the OpenPose format, utilizing MMPose~\cite{mmpose2020};
3) Depth Estimation, Surface Normal, and Edge Detection: given ImageNet and COCO images, we generate annotations in line with the protocols from~\cite{prismer}.
3) Style Transfer~\cite{brooks2022instructpix2pix}, De-rain~\cite{derain_zhang_2018}, De-noise~\cite{denoising_vincent}, Low Light Enhancement~\cite{wei2018deep}, and Stereo Datasets~\cite{kitti}: These are all represented as image pairs (e.g. input/output).
4) Colorization: We convert ImageNet images to greyscale, producing image pairs. 
5) Inpainting: The process involves randomly adding black-colored boxes in images to simulate corruption, resulting in image pairs.
For all the above annotation types, we can create visual sentences by concatenating 8 image pairs of the same annotation type into a 16-image visual sentence.  For datasets containing $k$ different annotations for the same image we use a different approach: for each set of $1+k$ images (input plus $k$ annotations), we randomly select $m$ elements, where $m \leq n+1 \leq 16$. These m-tuples are then concatenated to form visual sequences.

\vspace{1.3mm}
\noindent\textbf{Image sequences with annotations.} 
When converting annotated video data (VIPSeg~\cite{miao2022large}, Hand14K~\cite{fathi2011learning}, AVA~\cite{murray2012ava}, JHMDB~\cite{Jhuang:ICCV:2013}) to visual sentences, we apply two complementary strategies. The first is similar to how we treat image data with paired annotations: each visual sentence is constructed by concatenating frames with their annotations -- \{frame1,annot1,frame2,annot2,...\}. 
The second method involves grouping multiple frames followed by their corresponding annotations -- \{frame1,frame2,annot1,annot2,...\}.

We present a detailed summary of all the data sources, annotation type and data statistics of UVDv1 in the Appendix.

\section{Approach}

\begin{figure}
    \centering
    \includegraphics[width=\linewidth]{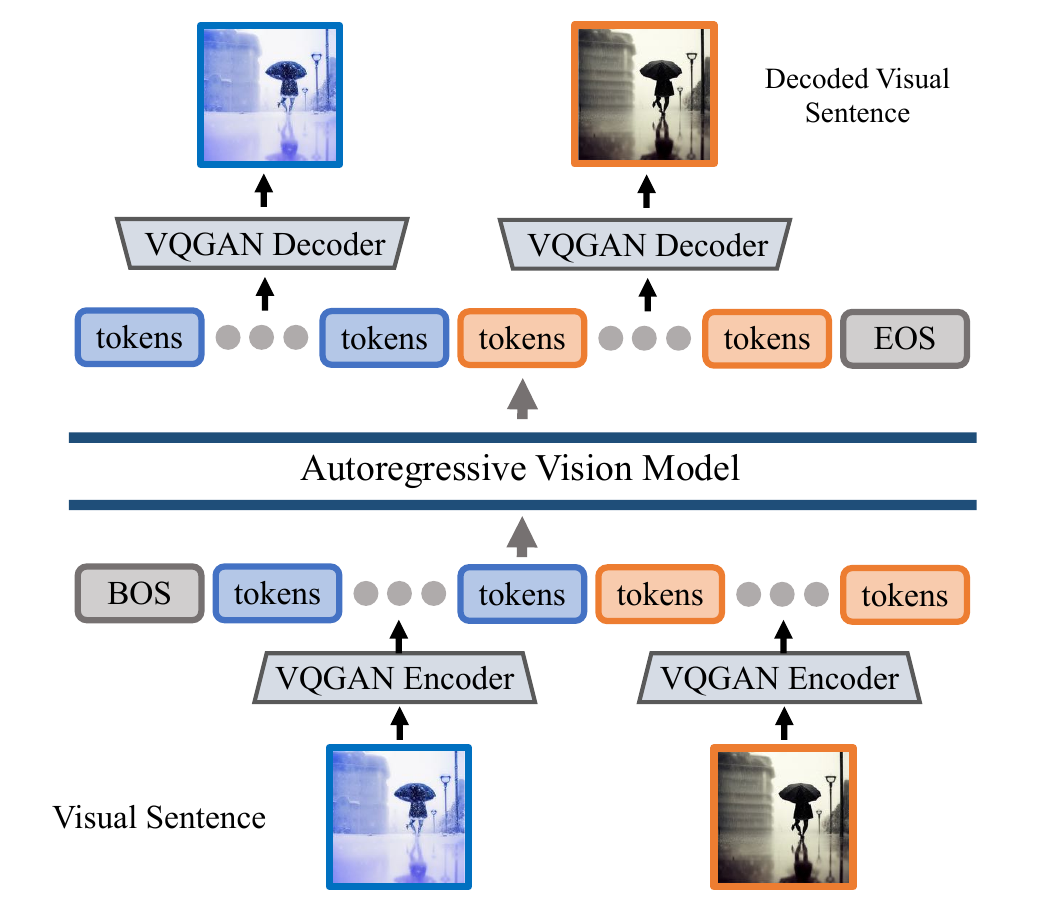}
    \vspace{-8mm}
    \caption{\small{\textbf{Architecture of \ours}. We first convert individual images from a visual sentence into discrete tokens using a VQGAN encoder. The resulting tokens from all images are then concatenated into a 1D sequence, and fed into an autoregressive Transformer model to predict the next token in the sequence. The predicted visual tokens are decoded into images using the VQGAN decoder.}}
    \label{fig:lvm_arch}
    \vspace{-6mm}
\end{figure}

In this section, we describe the design of our autoregressive Large Vision Model.  
Unlike text data, which naturally exhibits discrete sequential structure, it is not straightforward to model image pixels in visual sentences. 
In this work, we take a two-stage approach: 1) train a large visual tokenizer (which operates on individual images) to convert each image into a sequence of visual tokens; 2) train an autoregressive transformer model on visual sentences, each represented as a sequence of tokens. We summarize our approach in Figure~\ref{fig:lvm_arch}.

\subsection{Image Tokenization}

While the visual sentences exhibit a sequence structure between consecutive images, we don't have such natural sequence structure within an image. Therefore, in order to apply a transformer model to images, prior works typically do one of the following: either divide the image into patches in scan-line order, and treat that as a sequence~\cite{dosovitskiy2020image}, or use a pre-trained image tokenizer, such as VQVAE~\cite{van2017neural} or VQGAN~\cite{esser2021taming}, to cluster image features into a grid of 
discrete tokens, which, again, are turned into a sequence in scan-line order. 
We adopt the latter approach since the discrete categorical output from a model naturally forms a probabilistic distribution that one can easily sample from, enabling flexible conditional generation of new images within a visual sentence.

Specifically, we employ semantic tokens generated by a VQGAN model, a concept introduced by Esser et al~\cite{esser2021taming}. This framework consists of an encoding and a decoding mechanism, featuring a quantization layer that assigns input images to a sequence of discrete tokens from an established codebook. Our encoders and decoders are constructed purely with convolutional layers. The encoder is equipped with several downsampling modules to contract the spatial dimension of the input, whereas the decoder is fitted with an equivalent series of upsampling modules to restore the image to its initial size. For a given image, our VQGAN tokenizer produces  256 discrete tokens.

It is important to note that our tokenizer operates on individual images independently, rather than on the entire visual sentence at once. This independence allows us to decouple the tokenizer training from the downstream Transformer model so that the tokenizer can be trained on a dataset of single images without having to consider the distribution of visual sentences.

\vspace{1.3mm}
\noindent\textbf{Implementation Details:}
We adopt an off-the-shelf VQGAN architecture from \citet{chang2022maskgit}. We follow the exact  configuration in \citet{chang2022maskgit}, which uses a downsampling factor of \( f=16 \) and codebook size 8192. This means that for an image of size $256 \times 256$, our VQGAN tokenizer produces $16 \times 16 = 256$ tokens where each can take 8192 different values. We found that using the results of an ImageNet pre-trained tokenizer did not generalize well beyond ImageNet images.  Therefore, we train our own tokenizer on a 1.5B subset of the LAION 5B dataset~\cite{schuhmann2022laion5b}.

\subsection{Sequence Modeling of Visual Sentences}

After converting images into discrete tokens with VQGAN, we treat our visual sentence as a unified sequence by concatenating the discrete tokens from multiple images into a 1D sequence.  Importantly, all visual sentences are treated equally -- we do not make use of any special tokens to indicate particular tasks or formats.  We train a causal Transformer model with the next token prediction objective using a cross-entropy loss, similar to the standard approach for language models~\cite{gpt3}. Training the model the same way on all visual sentences enables the model to infer the relation between images from context instead of from task- or format-specific tokens. This gives the model an opportunity to generalize to other, unseen visual sentence structures.

\vspace{1.3mm}
\noindent\textbf{Implementation Details:}
After tokenizing each image in a visual sentence into 256 tokens, we concatenate them to form a 1D sequence of tokens. On top of the sequences of visual tokens, our Transformer model is virtually the same as an autoregressive language model, so we adopt the Transformer architecture of LLaMA~\cite{touvron2023llama}, a popular open-source language model with widely available implementations. We use a context length of 4096 tokens, which can fit 16 images under our VQGAN tokenizer. Similar to language models, we add a [BOS] (begin of sentence) token to the beginning of each visual sentence and an [EOS] (end of sentence) token to the end, and use sequence concatenation~\cite{chowdhery2022palm} during training time to improve efficiency. We train our model on our entire UVDv1 dataset ($420$ billion tokens) using one epoch (simple epoch training is standard in language models to avoid potential overfitting). We train 4 models with different numbers of parameters: 300 million, 600 million, 1 billion and 3 billion, following the same training configurations. We provide the detailed training hyperparameters in Appendix~\ref{app}.

\subsection{Inference by Visual Prompting}

Since the autoregressive Transformer in our model outputs a probability distribution of the next token conditioned on previous tokens, we can easily sample from this distribution to generate new visual tokens that complete a visual sentence. To use the model for downstream tasks, one can construct a partial visual sentence that defines a task at test time, and apply the model to generate the output. This is similar to in-context learning in language models~\cite{brown2020language} or visual prompting in computer vision~\citep{hertzmann2001image,bargandelsman2022visual}. 

\begin{figure}
    \centering
    \includegraphics[width=\linewidth]{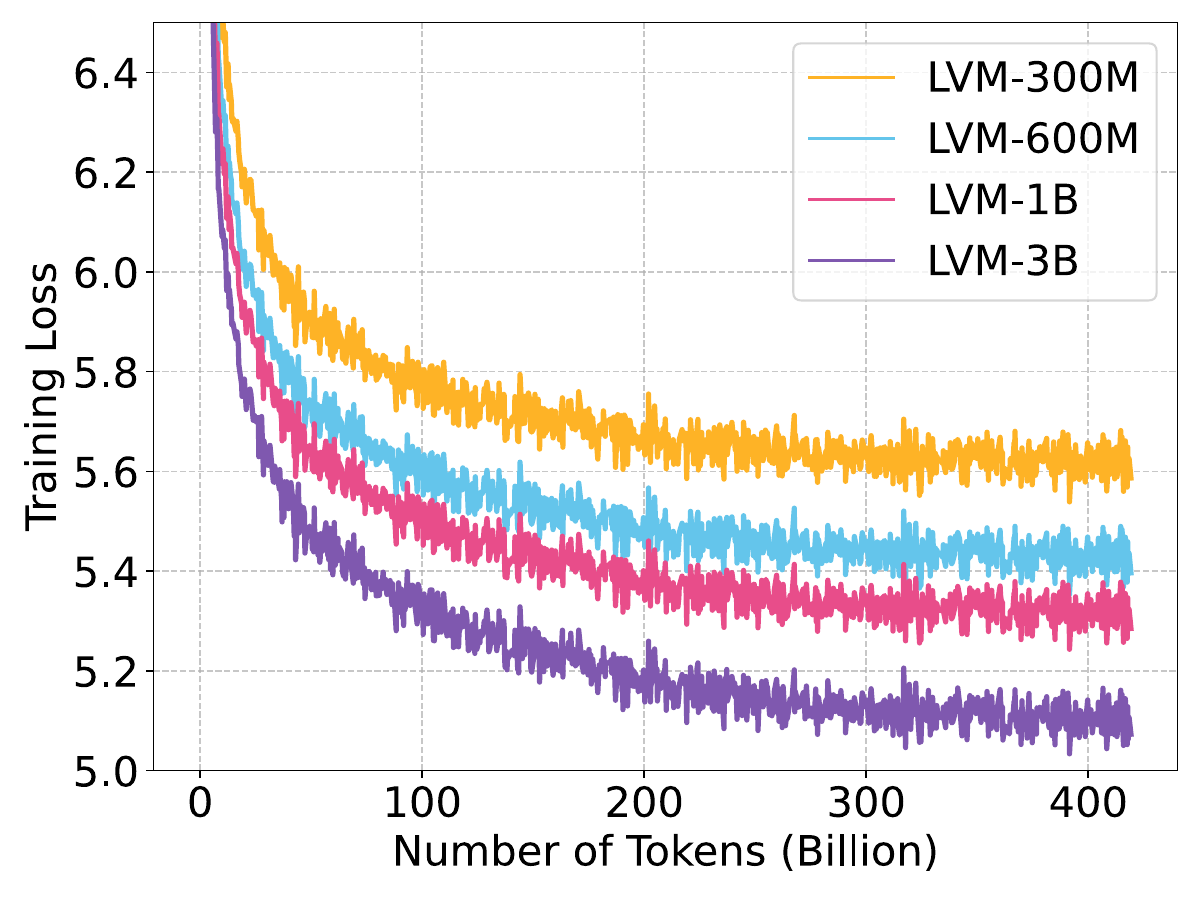}
    \vspace{-8mm}
    \caption{\small{\textbf{Training loss for the 300M, 600M, 1B, and 3B models.} All models are trained on 420B tokens, which correspond to 1.64B images. The training scales well with model sizes.}}
    \label{fig:scalability_training}
    \vspace{-3mm}
\end{figure}

\begin{figure}
    \centering
    \includegraphics[width=0.49\linewidth]{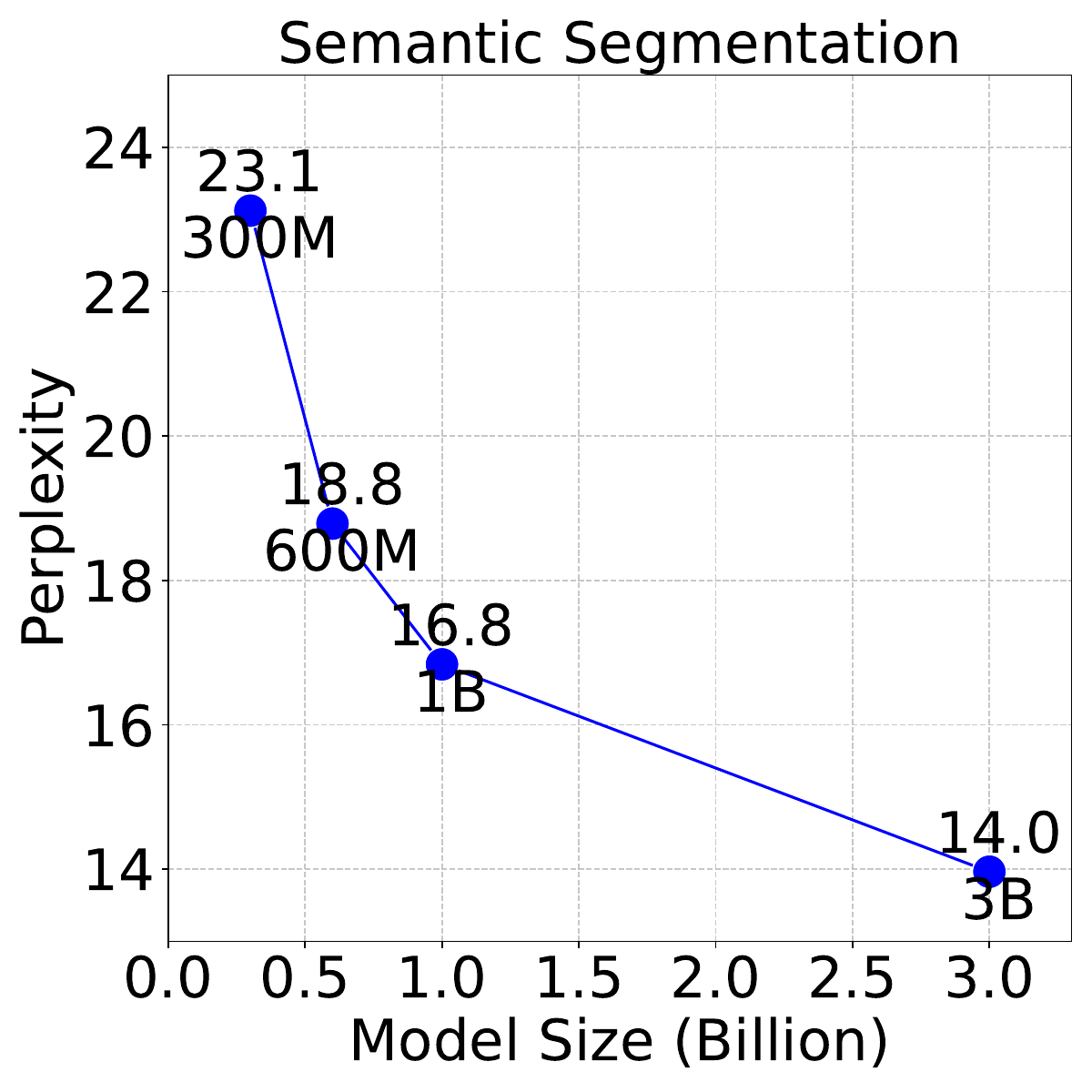}
    \includegraphics[width=0.49\linewidth]{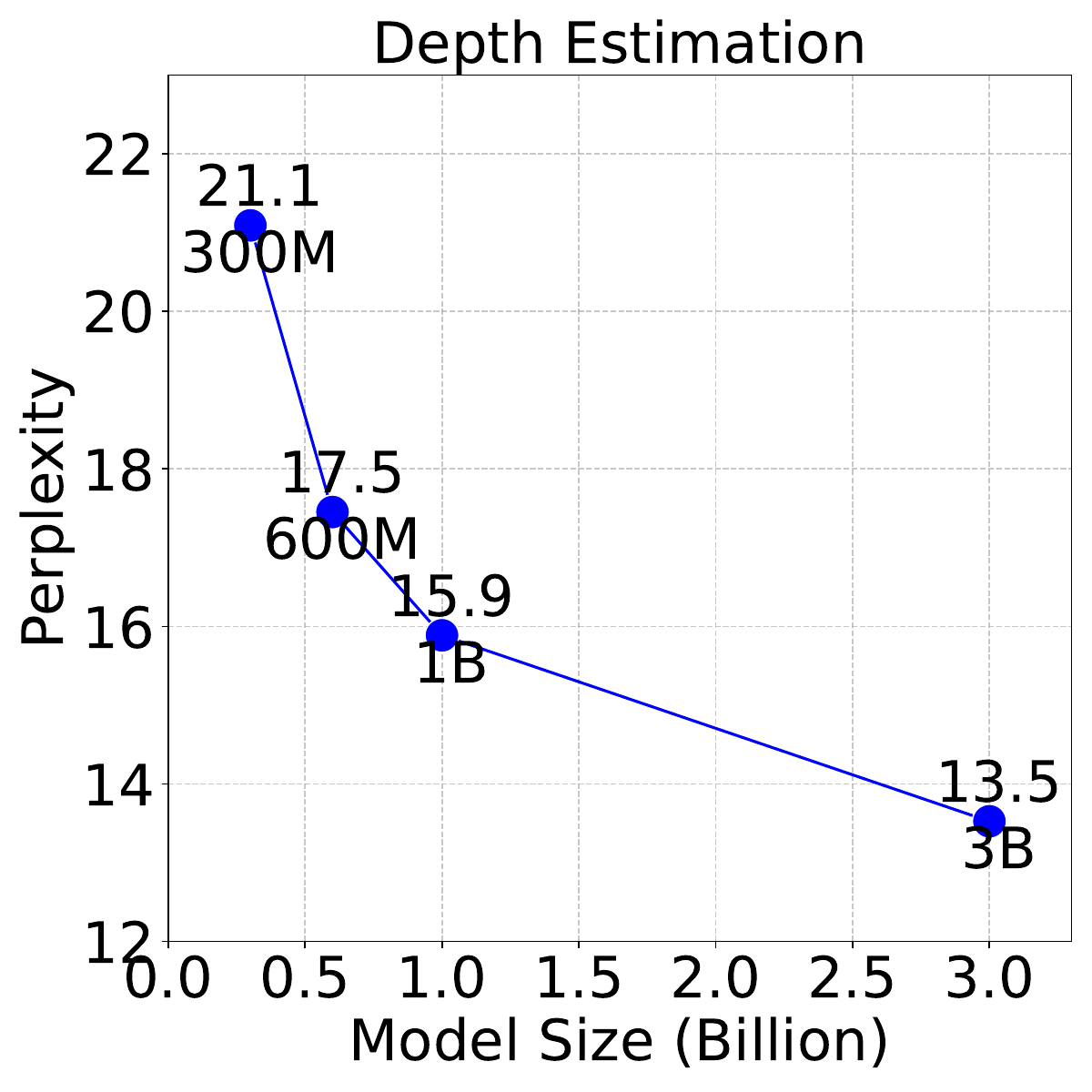}\\
    \includegraphics[width=0.49\linewidth]{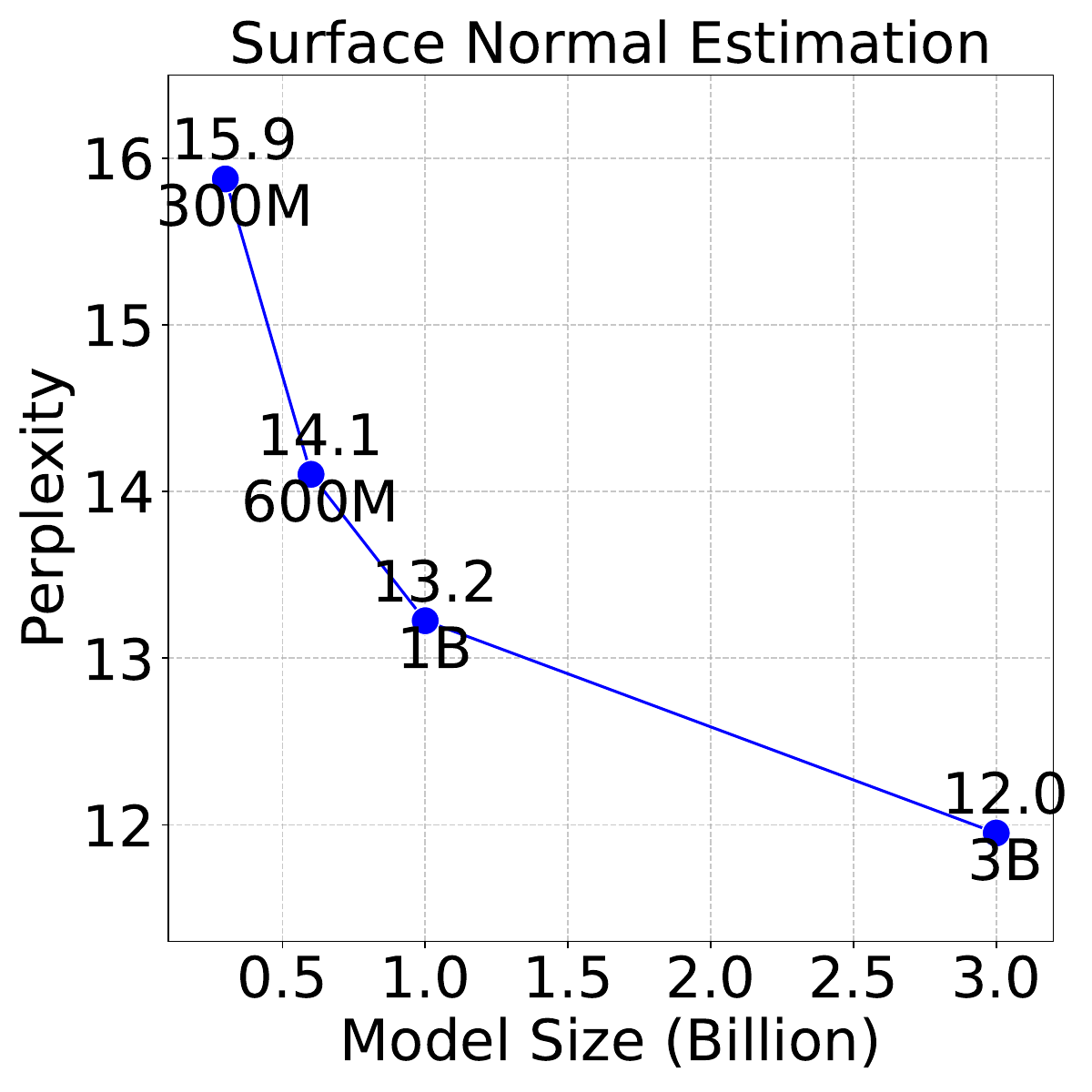}
    \includegraphics[width=0.49\linewidth]{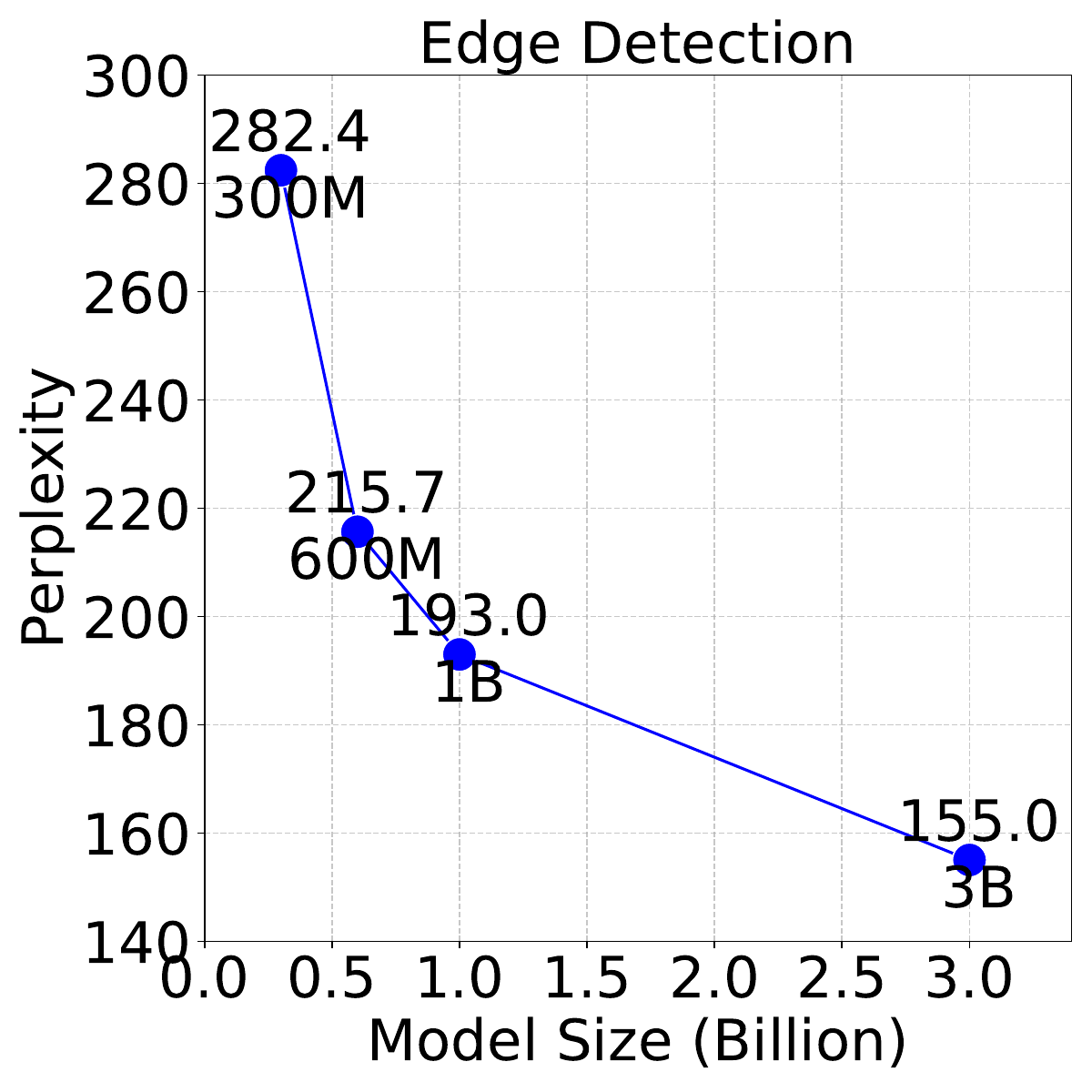}
    \vspace{-3mm}
    \caption{\small \textbf{Larger LVMs perform better on downstream tasks.} We evaluate \ours models of varying sizes on $4$ different downstream tasks, following the $5$ shot setting on the ImageNet validation set and report the perplexity. We find that perplexity decreases with larger models across all tasks, indicating the strong scalability of \ours.}
    \label{fig:scalability_task}
    \vspace{-4mm}
\end{figure}

\begin{figure*}
    \centering
    \includegraphics[width=0.97\linewidth]{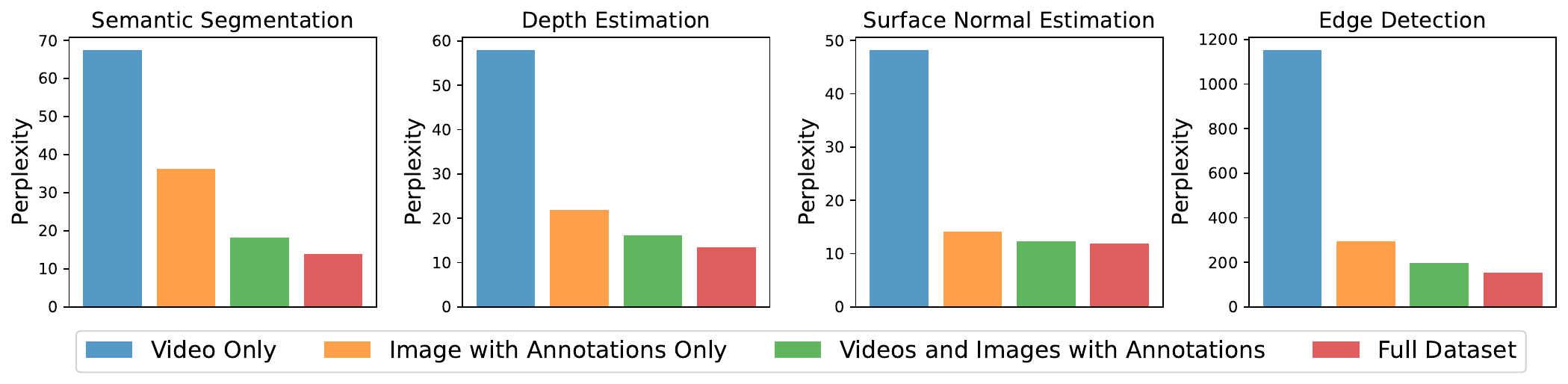}
    \vspace{-3mm}
    \caption{\small We evaluate the perplexity of 4 models trained on different subsets of our datasets on multiple tasks using the ImageNet validation set. All models are 3B parameters and all evaluations are conducted in the 5-shot setting. We can see that the model benefits from each single images, videos and annotations, demonstrating the importance of our training dataset diversity.}
    \vspace{-4mm}
    \label{fig:scalability_data}
\end{figure*}

\begin{figure*}
    \centering
    \includegraphics[width=\linewidth]{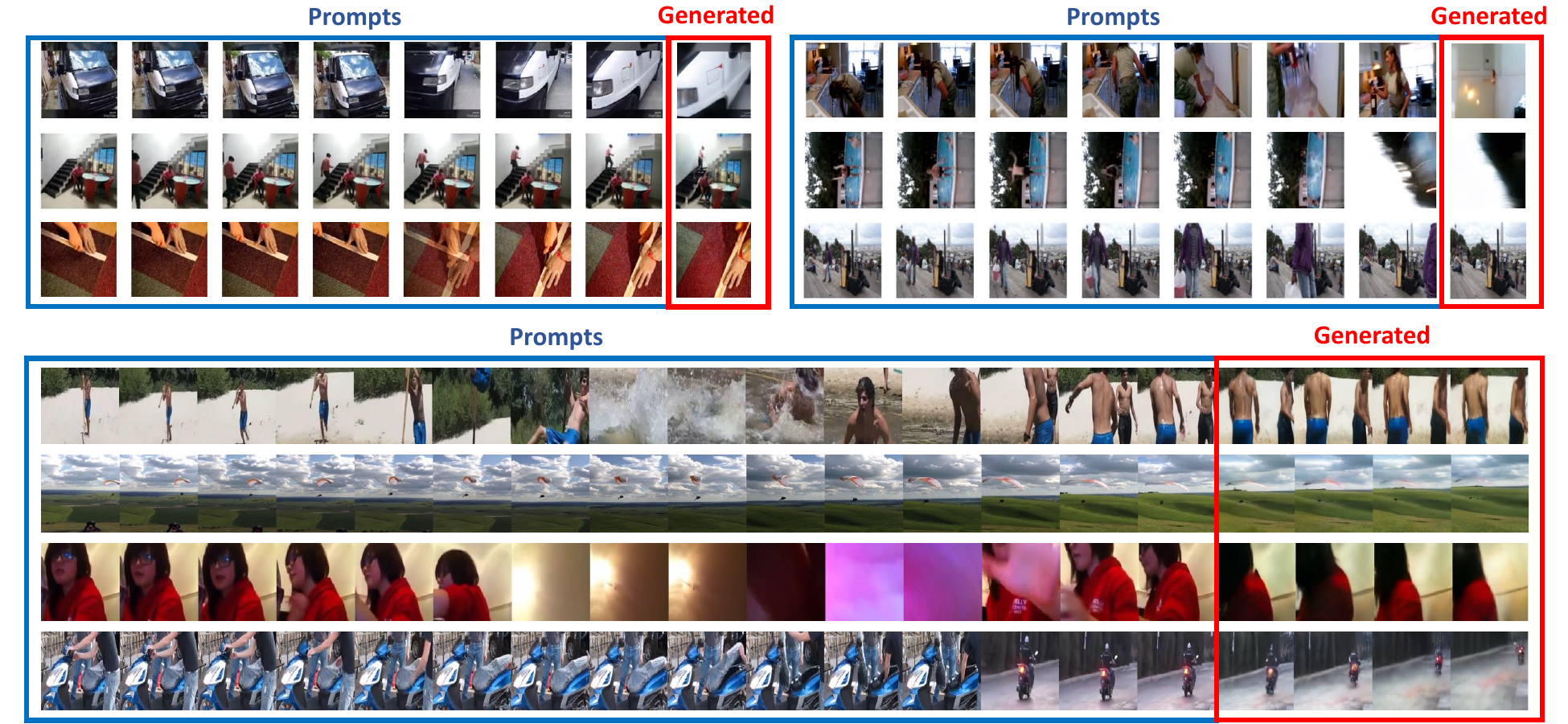}
    \vspace{-6mm}
    \caption{\small \textbf{Frame predictions.} \ours predicts the next frame (marked in red) given previous video frames as prompt. The results reveal that the \ours can predict the video frames while considering dynamic objects and camera motion.}
    \label{fig:video-generation}
    \vspace{-4mm}
\end{figure*}

\section{Experimental Results and Analysis} 
\label{sec:Experiment}
In this section, we evaluate the scaling abilities of our trained model, as well as its ability to understand and answer a range of diverse prompted tasks. 

\subsection{Scalability}

We investigate the scaling behavior of our model in terms of the training loss and downstream task performance as we increase the model size as well as the number of tokens seen during training.

\vspace{1.3mm}
\noindent\textbf{Training loss.} We first inspect the training loss of \ours with different parameter sizes, which we present in Figure~\ref{fig:scalability_training}. Since all our models are trained for only one epoch on the dataset, the model sees a given data sample just once, and therefore the training loss at any point during training is very similar to the validation loss. One can observe that as training progresses: 1) the training loss (perplexity) of the models, regardless of their size, continues to decrease; 2) as we increase the size of the model (parameter count), the loss decreases faster.  These observations indicate that \ours shows strong scalability behavior with both larger models and more data.

\vspace{1.3mm}
\noindent\textbf{Scalability on downstream benchmarks.} While the \ours  overall loss scales well during training, there is no guarantee that the better overall model would also perform better on a given specific downstream task. Therefore, we evaluate different sizes of models on 4 downstream tasks: semantic segmentation, depth estimation, surface normal estimation, and edge detection. We evaluate these tasks on the ImageNet validation set and generate all the annotations using the corresponding method described in Sec.~3. For each task, we give 5 pairs consisting of the inputs and corresponding ground-truth annotations as well as the query image as input prompt and evaluate the perplexity of the ground-truth annotation under our model's prediction of the next $256$ tokens (one image). We report the results in Figure~\ref{fig:scalability_task}. We see that larger models indeed attain lower perplexity across all tasks, showcasing that our scalable overall performance does transfer to a range of downstream tasks. 

\vspace{1.3mm}
\noindent\textbf{Dataset ablation.}
While \ours attains better performance with larger models and more data, it is natural to ask whether each data component we collect in UVDv1 helps. To answer this question, we conduct an ablation study on our dataset by training several 3B models on subsets of our dataset, and compare their performances on downstream tasks. We use the same 4 downstream tasks and settings as before and present the results in Figure~\ref{fig:scalability_data}. We observe that each data component contributes positively to the downstream tasks. \ours not only benefits from larger data, but also improves with more diversity in the dataset, which includes both annotated and unsupervised image and video data.

\subsection{Sequential Prompting}

We begin with the most intuitive and straightforward approach to visually prompt the LVM: sequential reasoning.  Here the prompt construction is very simple: we present the model with a sequence of 7 images and ask it to predict the next image ($256$ tokens).

\begin{figure}
    \centering
    \includegraphics[width=0.7\linewidth]{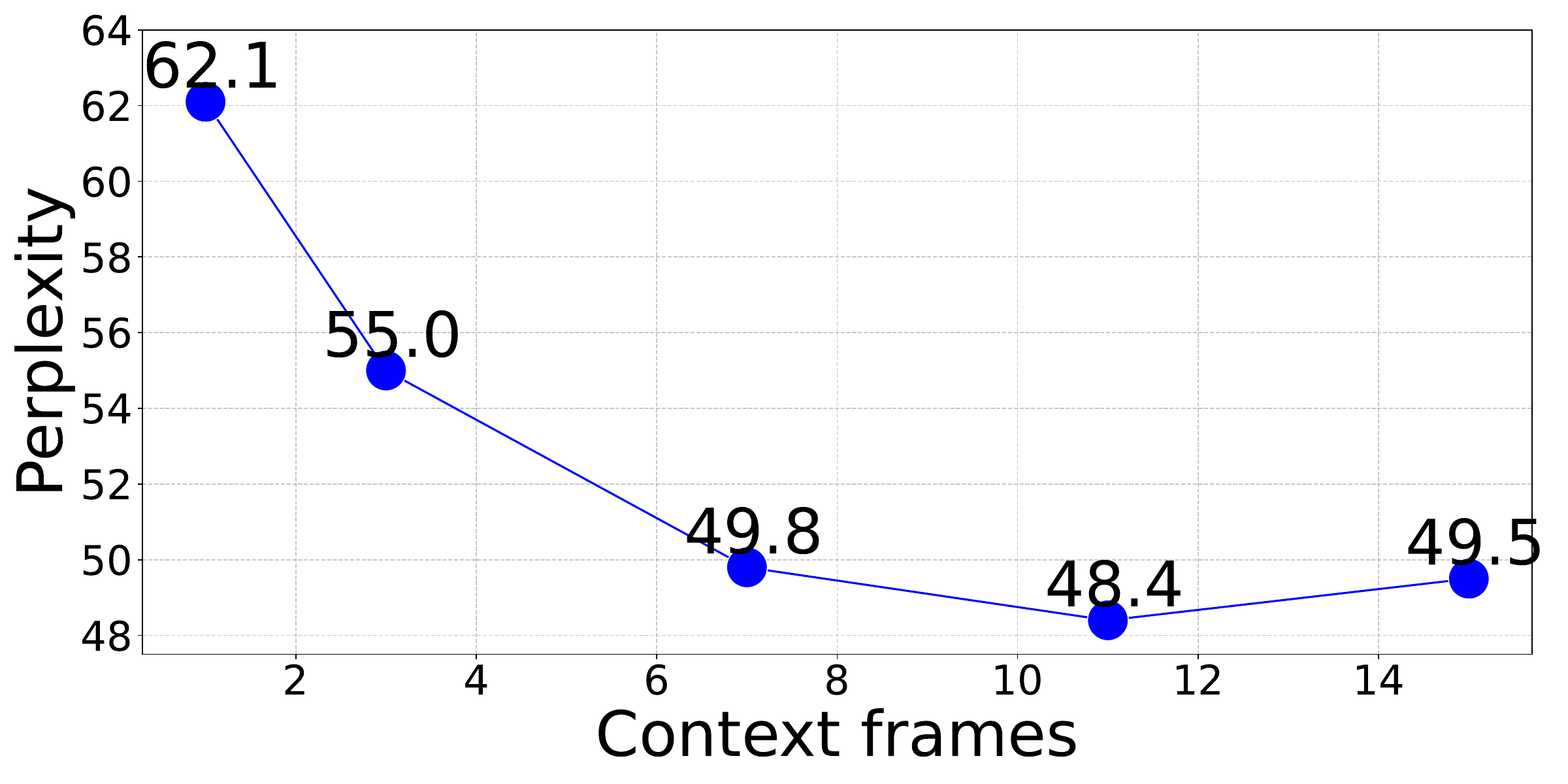}
    \vspace{-4mm}
    \caption{Longer context helps model understand better.}
    \label{fig:context_length}
    \vspace{-6mm}
\end{figure}

\begin{figure*}
    \centering
    \includegraphics[width=\linewidth]{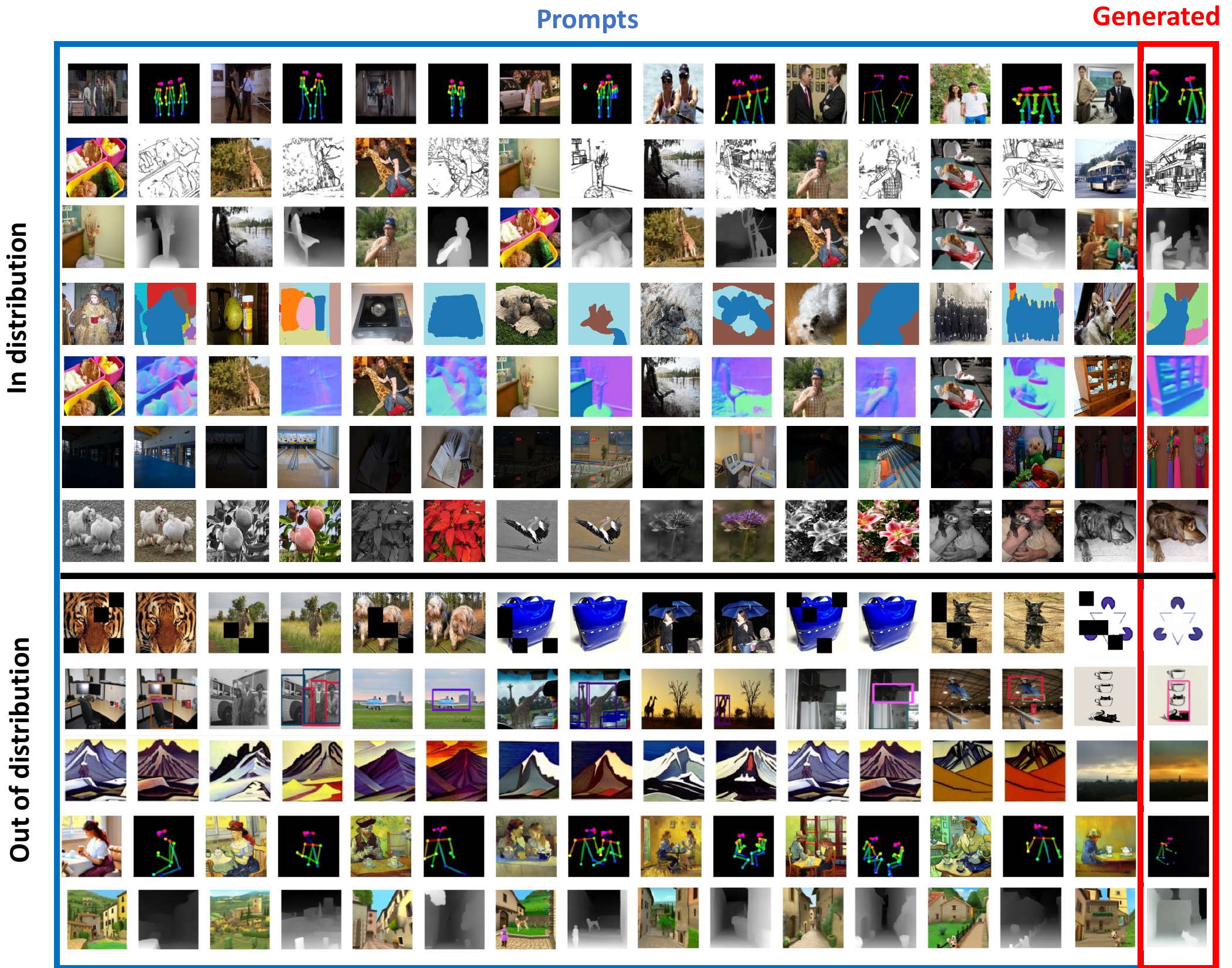}
    \vspace{-6mm}
    \caption{\small \textbf{In and out of distribution prompting examples.} Every row is a prompt that contains a sequence of images interleaved with annotations, followed by a query. The last image is predicted by the model (marked in red). The last 5 rows show examples where the query image is out of distribution (painting, sketch, etc) for the task it was trained for. 
    }
    \label{fig:image_tasks}
    \vspace{-5mm}
\end{figure*}

\vspace{1.3mm}
\noindent\textbf{Video frame prediction.} The most direct task for sequential prompting is video prediction.  Figure~\ref{fig:video-generation} presents several next frame prediction examples, prompted by sequences from the Kinetics-700 validation set.  At the top, 7 frame prompts (blue border) are followed by the predicted frame (red border). We observe a certain degree of inferential ability regarding spatial positioning, viewpoint and object understanding. Perplexity of prediction on Kinetics val set is 49.8.  The last 4 rows show predictions with longer context (15 frames) and a longer prediction (4 frames).  See \cref{fig:k700_1,fig:k700_2,fig:k700_3,fig:k700_4,fig:k700_5,fig:k700_6} in the Appendix for many more examples.  

\vspace{1.3mm}
\noindent\textbf{Rotation and Category prediction.} The same type of simple sequential prompting can be used in other ways as well.  For example, Figure~\ref{fig:3d}  shows how prompting the model with a sequence of 3D rotations of a synthetic object around an arbitrary axis allows it to predict further rotation.  Or we can think of a list of items of a given category as a sequence and predict other ideas in that same category, as shown in Figure~\ref{fig:sketch}.  Note that, while the system was trained on groups of images from the same ImageNet category, here the prompt consists of sketches, which have not been seen in any annotated data.

\noindent\textbf{Context length analysis.} Next we ask how much temporal context is required to accurately predict the subsequent frame? We assessed the model's frame generation perplexity when prompted with a context of varying lengths (1 to 15 frames).  As  Figure~\ref{fig:context_length} shows, on the Kinetics-700 val set, we see a clear improvement in perplexity from 1 to 11 frames after which it stabilizes (from $62.1 \rightarrow 48.4$).

\begin{figure*}
    \centering
    \includegraphics[width=\linewidth]{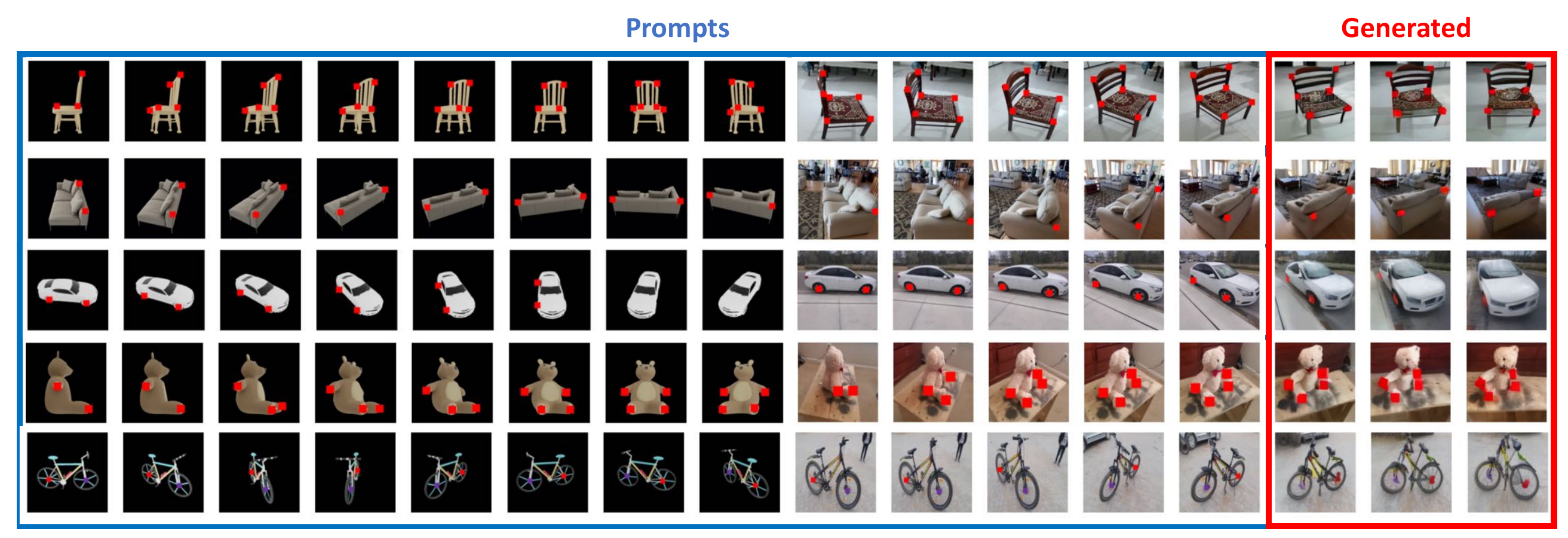}
    \vspace{-7mm}
    \caption{\textbf{Task Compositionality}. Examples of prompts that combine two different tasks -- object rotation and keypoint tracking. }
    \label{fig:complex_tasks}
    \vspace{-4mm}
\end{figure*}

\begin{figure}[t!]
    \centering
    \includegraphics[width=\linewidth]{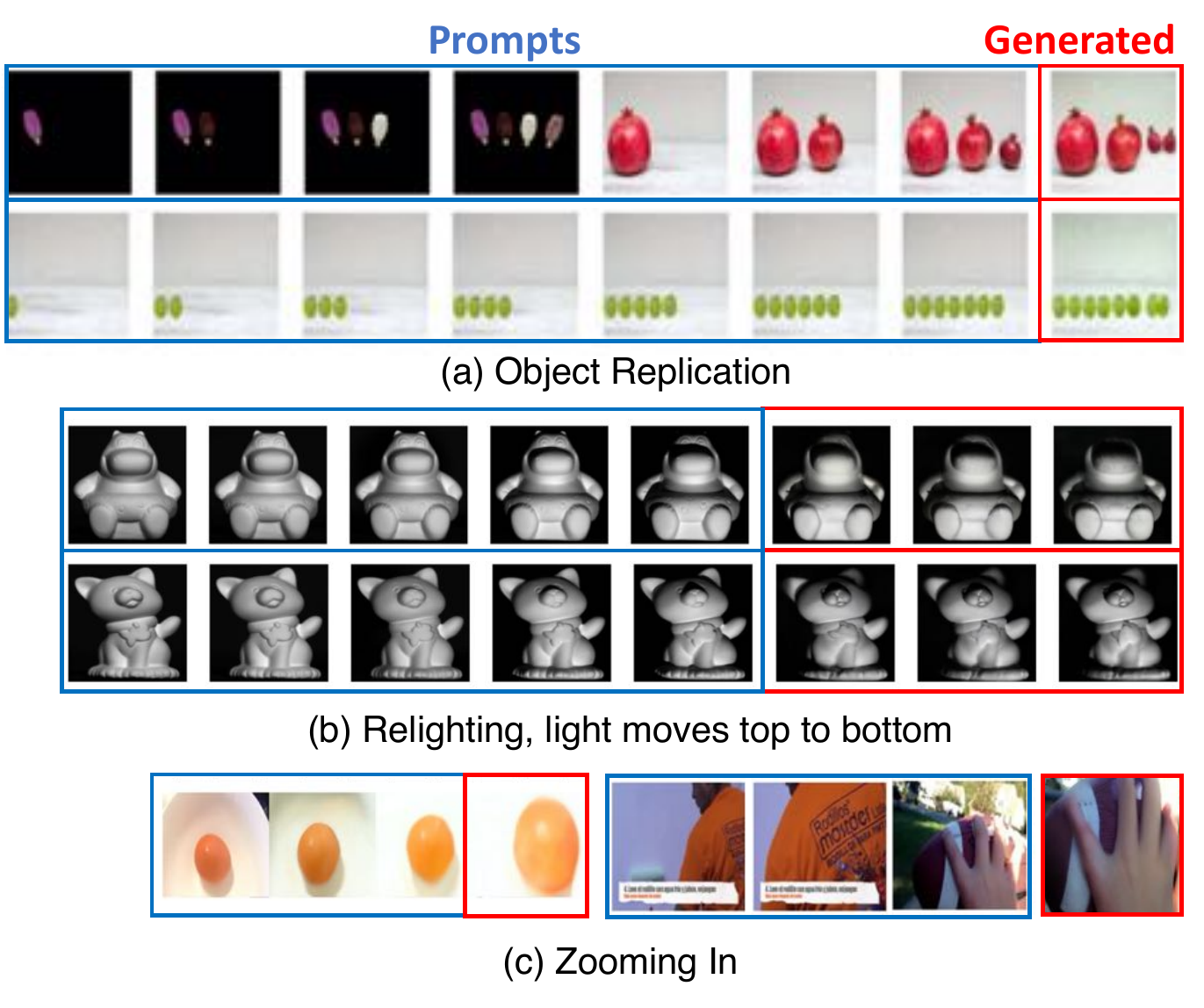}
    \vspace{-9mm}
    \caption{\textbf{Miscellaneous Prompting}. A variety of simple vision tasks, such as object replication (top), relighting (middle), and zooming in (bottom), can be simply specified via a suitably chosen visual sentence prompt that expresses the task to the LVM.}
    \label{fig:misc}
     \vspace{-6mm}
\end{figure}
\subsection{Analogy Prompting}
\vspace{-3mm}
Our study progresses by evaluating a more complex prompting structure, which we call `Analogy Prompting'. This method challenges the model to comprehend analogies of arbitrary length and complexity, thereby testing its advanced interpretative abilities.

\noindent\textbf{Qualitative Results. }
Figure~\ref{fig:image_tasks} shows a sampling of qualitative results with analogy prompting on a number of tasks.  The prompts consist of a sequence of 14 images giving examples of various tasks, followed by a 15th query image. Given each prompt, the next image predicted is the result.  The top part of the figure shows several example prompts defining tasks that were part of the training set (but these actual images were never seen at training). The bottom part of the figure demonstrates generalization to tasks never shown at training. See the Appendix for many more qualitative examples.

\noindent\textbf{Unseen Tasks and Dataset.} We present the results for keypoint detection on Pascal 3D+~\cite{pascal3d}, evaluated using the standard Percentage of Correct Keypoints (PCK) metric with a of threshold 0.1. Remarkably, \ours achieves a PCK of 81.2 without training on this dataset, demonstrating impressive generalization capabilities. In comparison, we show some existing task-specific model: StackedHourglass~\cite{newell2016stacked} scores 68.0 PCK, MSS-Net~\cite{ke2018multi} achieves 68.9 PCK, and StarMap~\cite{zhou2018starmap} registers 78.6 PCK.

\vspace{1.3mm}
\noindent\textbf{Comparison with Visual Prompting.}
The closest approach to ours that also allows for defining arbitrary tasks is Visual Prompting~\cite{bargandelsman2022visual}. 
In Table~\ref{tab:main_table}, we compare various visual prompting models on few-shot segmentation, object detection, and colorization tasks. Note that our sequential LVM beats previous approaches on almost all tasks.

\vspace{1.3mm}
\noindent\textbf{Task Compositing.} 
Figure~\ref{fig:complex_tasks} demonstrates compositing several tasks together within a single prompt.  Here, we demonstrate the rotation task together with the novel keypoint correspondence task and ask the model to continue the pattern. The model is able to successfully combine these two at test ti   me, demonstrating some degree of compositionality. 

\begin{figure}
    \centering
    \includegraphics[width=\linewidth]{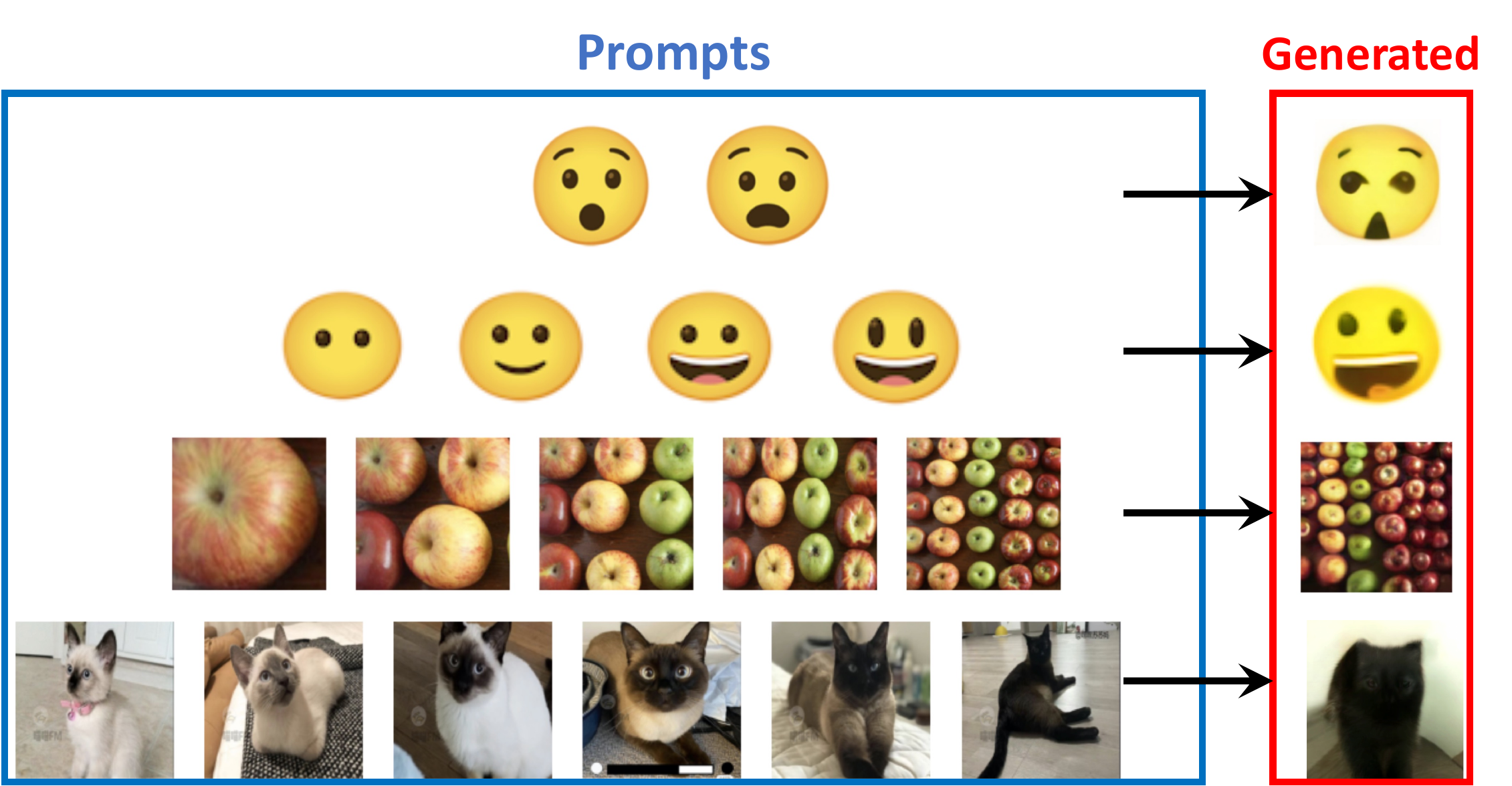}
    \vspace{-5mm}
    \caption{\textbf{What comes next?} Tasks that are not always easily describable in language.}
    \label{fig:guess}
\end{figure}

\begin{table}
  \centering
\resizebox{\linewidth}{!}{

  \begin{tabular}{l|rrrr|rrrr|cc}
    \toprule
    Model &  \multicolumn{4}{c}{Foreground Segmentation~$\uparrow$} & \multicolumn{4}{c}{Single Object Detection~$\uparrow$} & \multicolumn{2}{c}{Colorization~$\downarrow$} \\
         & Split 0 & Split 1 & Split 2 & Split 3 & Split 1 & Split 2 & Split 3 & Split 4 & MSE & LPIPS \\
    \midrule
    MAE (IN-1k) & 1.92 & 6.76 & 3.85 & 4.57 & 1.37 & 1.98 & 1.62 & 1.62 & 1.13 & 0.87\\
    MAE-VQGAN (IN-1k) & 2.22 & 7.07 & 5.48 & 6.28 & 3.34 & 3.21 & 2.80 & 2.80 & 3.31 & 0.75 \\ 
    \midrule
    MAE (CVF)       & 17.42 & 25.70 & 18.64 & 16.53 & 5.49 & 4.98 & 5.24 & 5.84 & \textbf{0.43} &  {0.55}\\ 
    MAE-VQGAN (CVF)  & {27.83} & {30.44} & {26.15} & {24.25} & {24.19} & {25.20} & {25.36} & {25.23} & 0.67  & \textbf{{0.40}} \\ \hline
    
    Ours & \textbf{48.94} & \textbf{ 51.29} & \textbf{47.66} &\textbf{ 50.82} &\textbf{48.25}  & \textbf{49.60} & \textbf{50.08 }& \textbf{48.92} &  {0.51} & {0.46} \\ %
\bottomrule
  \end{tabular}}
   \vspace{-2mm}
    \caption{\textbf{Comparison with Visual Prompting~\cite{bargandelsman2022visual}.} For Foreground Segmentation and Single Object Detection, we report the \textit{mIOU} score. For Colorization, we report the \textit{MSE} and \textit{LPIPS}.}
    \vspace{-4mm}
  \label{tab:main_table}
\end{table}

\begin{figure*}[!t]
    \centering
\includegraphics[width=0.7\linewidth]{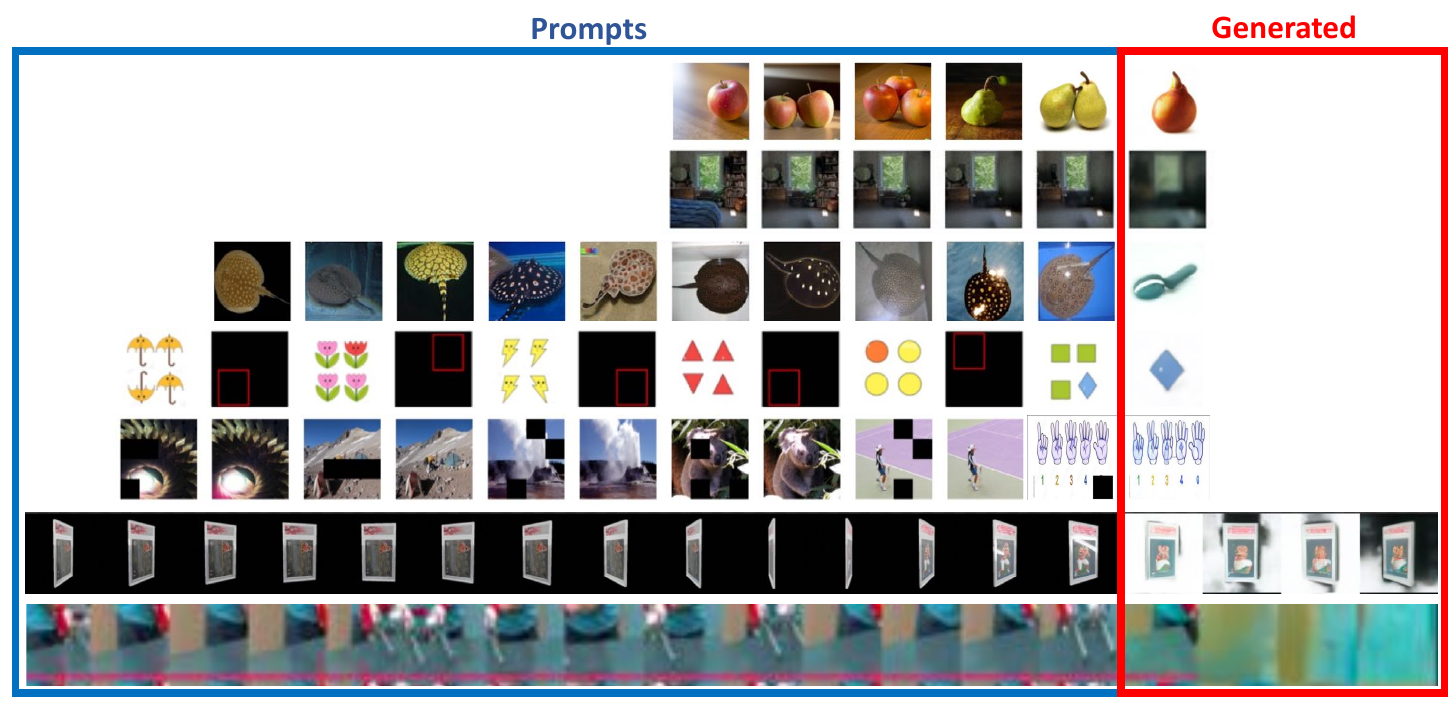}
\vspace{-3mm}    
\caption{\textbf{Failure Cases.} This figure illustrates seven examples of failure cases: (1) Task confusion: the model interprets the counting task as style transfer, resulting in a pear-shaped apple. (2) Task entanglement: the model perceives the prompt as high frequencies from the entire image, rather than just from some parts of image (objects). (3) Wrong instance: the model correctly identifies rotation but mistakenly generates a brush instead of a manta. (4) Outlier detection versus generation: rather than detecting an outlier, the model directly generates it. (5) Digits: the model fails to represents digit sequences. (6) Tokenizer: poor performance with some out-of-distribution synthetic data generation (wrong background). (7) Sequence degeneration: sometimes frame predictions go astray.}
    \vspace{-4mm}
    \label{fig:failure}
\end{figure*}

\begin{figure*}[t!]
    \centering
    \includegraphics[width=0.95\linewidth]{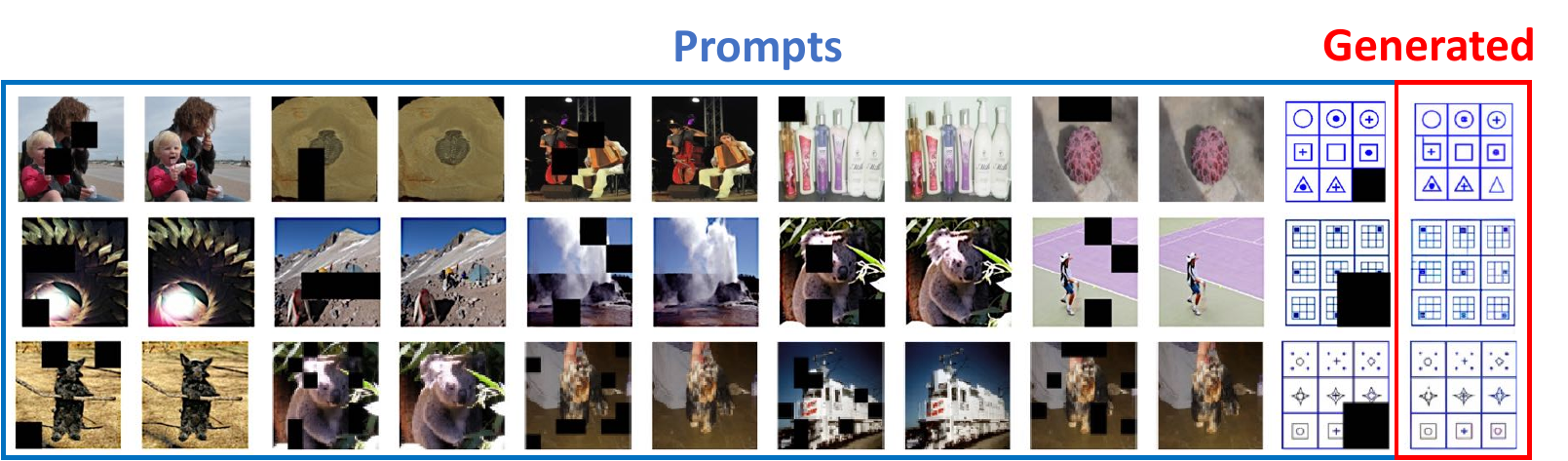}
        \vspace{-3mm}
    \caption{\small \textbf{``Sparks of AGI?" \includegraphics[width=0.03\linewidth]{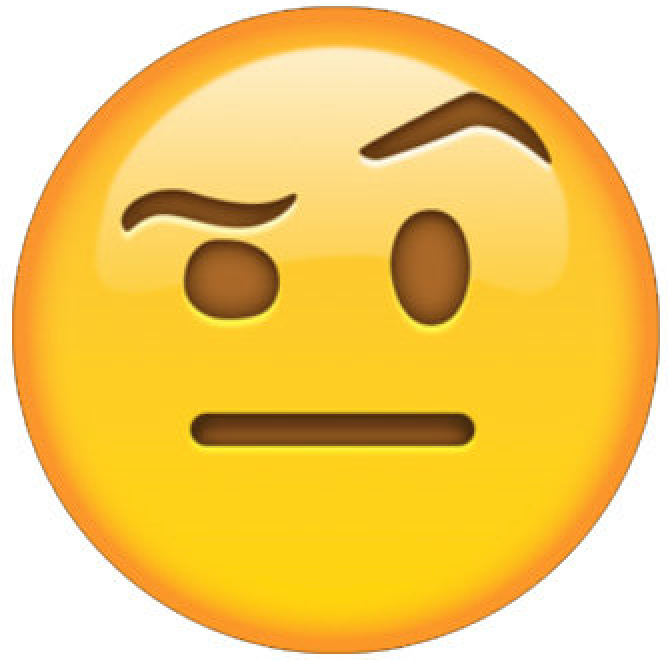}} We prompt LVM with a masked reasoning visual sentence to infer the solution for non-verbal reasoning questions which are prevalent in IQ tests (masked image, second from the right). We find that the model often infers and applies the abstract visual pattern correctly. So, we graciously hand over to you, our gentle reader, the task of pondering whether our modest LVM also exhibits the much-vaunted ‘Sparks of AGI’.}
    \label{fig:sparks}
          \vspace{-5mm}
\end{figure*}

\subsection{Miscellaneous Prompts} 
Here we try to see how far we can push our model by offering it various prompts it has not seen before. Figure~\ref{fig:misc} shows a few such prompts that happened to work reasonably well.  Figure~\ref{fig:guess} shows some prompts which are not easily describable by words -- these are the type of tasks where LVMs might eventually outshine LLMs. 

In Figure~\ref{fig:sparks}, we show initial qualitative results on a typical visual reasoning question as found on non-verbal human IQ tests (Raven's Progressive Matrices~\cite{raven1941standardization}). With considerable squinting, one could imagine the LVM having a latent ability for grasping abstract visual patterns and applying the grasped pattern to extrapolate the shown visual sequence.  This exciting result warrants further study. 

\section{Limitations} 

Figure~\ref{fig:failure} shows some typical failure cases of the current model. One common element, the use of visual prompt to define a task is often under-constrained (more so than in language, since images are very high-dimensional), or the requested task might be beyond the capabilities of the current system.  Other, more mundane failures involve issues with the tokenizer and lack of high-quality video training data.   

Limited computing resources placed severe constraints that prevented us from exploring a range of intriguing problems, including the impact of different data sets and detailed ablation studies. It is important to note that, despite this being one of the biggest vision models to date, it is still rather small in comparison with modern Large Language Models.  Therefore, the question of emergence and true generalization in Large Vision Models remains wide open and ripe for further study.  

\vspace{-5mm}
\paragraph{Acknowledgements:}  We are grateful to many friends and colleagues for the discussions and comments on this work, including Yossi Gandelsman, Aleksander Holynski, Angjoo Kanazawa, Qianqian Wang, Sophia Koepke, Quoc Le, Chen Liang, Ekin D. Cubuk, Assaf Shocher, Amir Zamir, Carl Vondrick, Ludwig Schmidt, Aviral Kumar, Xiaolong Wang, Yonglong Tian, Miki Rubinstein and Dilip Krishnan. 
The work has been supported, in part, by ONR MURI N0001 4-22-1-2773, N00014-21-1-2801, and N00014-21-1-2812, ERC HOLI, Apple Graduate Fellowship to Yutong Bai, and compute donation via the Google TPU Research Cloud.

\newpage

\clearpage
{
    \small
    \bibliographystyle{ieeenat_fullname}
    \bibliography{main}
}

\appendix
\clearpage
\appendix
\setcounter{page}{1}

\section*{Appendix Overview}
\label{app}

This supplementary document complements the main manuscript by providing detailed insights and additional support. It is structured as follows:

     \paragraph{Appendix A: Large Vision Models (LVMs) Detailed Overview} – Explores the specifics of LVMs used in our study, including model sizes, architectural details, and optimization hyperparameters.

     \paragraph{Appendix B: Unified Vision Dataset (UVD) In-Depth Analysis} – Provides a comprehensive examination of UVD, discussing its composition, data distribution and more details.

     \paragraph{Appendix C: Additional Results} – Offers extended results and visual evidence for our study, including supplementary figures and quantitative assessments.

\section{Approach: Large Vision Models (LVMs)}

\subsection{Model Architectures.}
As stated before, we use the Transformer variant of LLaMA~\cite{touvron2023llama} as our model architecture. To form different model sizes, we vary the hiddmen dimension, MLP intermediate dimension, number of heads and number of layers. We present the details in Table~\ref{tab:arch_config}. For the rests of the hyperparameters, we keep them the same as the standard LLaMA model.
\begin{table}[!h]
\centering
\footnotesize

\begin{tabular}{l|c|c|c|c}
\toprule
         & \textbf{hidden dim} & \textbf{MLP dim} & \textbf{heads} & \textbf{layers} \\
\midrule
LVM-300M & 1024      & 2688    &  8       & 22 \\
LVM-600M & 1536      & 4096    &  16      & 22\\
LVM-1B   & 2048      & 5504    &  16      & 22\\
LVM-3B   & 3200      & 8640    &  32      & 26 \\
\bottomrule
\end{tabular}
\caption{Model architecture configurations of LVMs.}
\label{tab:arch_config}
\end{table}
\subsection{Training and optimizer details.}
Folling the LLaMA~\cite{touvron2023llama} model, we use the AdamW optimizer to train our models. We use the same optimizer hyperparameters for all our models, and we present them in Table~\ref{tab:opt_hparams}. All our models are trained on TPU-v3 pods on Google Cloud. Our largest model, LVM-3B, takes around 14 days to train on a v3-512 TPU pod.

\begin{table}[!h]
    \centering
    \footnotesize
    \begin{tabular}{l | c}
        \toprule
        \textbf{Hyperparameter} & \textbf{Value} \\
        \midrule
        Learning rate schedule & linear warmup and cosine decay \\
        Base learning rate & 1.5e-4 \\
        Final learning rate & 1.5e-5 \\
        Warmup steps & 2000 \\
        Decay steps & 144000 \\
        Weight decay & 0.1 \\
        Optimizer & AdamW \\
        Optimizer momentum & $\beta_1 = 0.9, \beta_2 = 0.95$ \\
        Batch size & 2097152 tokens \\
        Context length & 4096 tokens \\
        \bottomrule
    \end{tabular}
    \caption{Hyperparameters for pre-training \ours}
    \label{tab:opt_hparams}
\end{table}

\section{Unified Vision Dataset (UVD) Details}

\subsection{Overview} 
The Unified Vision Dataset (UVD) represents an extensive compilation of visual data spanning a wide array of domains and annotation types. It integrates a diverse set of datasets, each contributing unique characteristics and annotations, thereby creating a rich resource for various vision-related tasks. The following Table~\ref{tab:UVDv1} provides a detailed overview of UVD, categorizing the datasets into specific groups based on their content type and annotation features. This categorization includes unpaired image data, images with annotations, videos, videos with annotations, and synthetic 3D views. Each dataset within these categories is listed with its corresponding token count, annotation type, and annotation source, offering a comprehensive perspective of the UVD's structure and composition.

\begin{table*}
\centering
\footnotesize

\begin{tabular}{|l|c|l|l|}
\toprule
\textbf{Dataset} & \textbf{Tokens (Millions)} & \textbf{Annotation Type}  & \textbf{Annotation Source} \\ 
\midrule
\multicolumn{4}{|c|}{\textbf{Unpaired Image Data}} \\ 
\midrule
LAION 5B~\cite{schuhmann2022laion5b} (1.5B images subset) & 380690 & - & - \\ 
\midrule
\multicolumn{4}{|c|}{\textbf{Images with Annotations}} \\ 
\midrule
ImageNet 1K~\cite{deng2009imagenet}          & 1317.40 & Image Classification                           & Ground Truth                        \\ 
COCO~\cite{lin2014microsoft}                 & 363 & Object Detection                               & MMDetection~\cite{mmdetection}      \\ 
ADE 20K~\cite{zhou2019semantic}, Cityscapes~\cite{Cordts2016Cityscapes} & 66.88 & Semantic Segmentation                          & Ground Truth                        \\ 
COCO~\cite{lin2014microsoft}, ImageNet 1K~\cite{deng2009imagenet} & 2078.06 & Semantic Segmentation                          & Mask2Former~\cite{cheng2021mask2former} \\ 
COCO~\cite{lin2014microsoft}, lvmhp~\cite{li2017multiple}, mpii~\cite{andriluka14cvpr}, Unite~\cite{lassner2017unite} & 950.79 & Human Pose  & MMPose\cite{mmpose2020}             \\ 
COCO~\cite{lin2014microsoft}, ImageNet 1K~\cite{deng2009imagenet} & 1623.85 & Depth Map Image    & DPT~\cite{ranftl2021vision}         \\ 
Subset of InstructPix2Pix~\cite{geiger2012we} & 415.46 & Style Transfer                                 & InstructPix2Pix~\cite{geiger2012we} \\ 
COCO\cite{lin2014microsoft}, ImageNet 1K\cite{deng2009imagenet} & 1623.85 & Surface Normal Image                                 & NLL-AngMF~\cite{bae2021estimating}  \\ 
COCO~\cite{lin2014microsoft}, ImageNet 1K~\cite{deng2009imagenet} & 1623.85 & Edge Detection                                  & DexiNed~\cite{xsoria2020dexined}    \\ 
DID-MDN~\cite{derain_zhang_2018}             & 35.06 & Rainy and Clean Image Pairs                                         & Ground Truth                        \\ 
SIDD~\cite{abdelhamed2018high}             & 245.76 & Denoised Image                                         & Ground Truth                        \\ 
LOL\cite{wei2018deep}                        & 0.458 & Light Enhanced Image                                      & Ground Truth                        \\ 
ImageNet 1K~\cite{deng2009imagenet}          & 1321.07 & Grayscale and Colorized Image Pairs                                   & Ground Truth                        \\ 
ImageNet 1K~\cite{deng2009imagenet}          & 1321.07 & Inpainting                                     & Ground Truth                        \\ 
Kitti~\cite{geiger2012we}                    & 9.21 & Stereo                                         & Ground Truth                        \\ 
\midrule
\multicolumn{4}{|c|}{\textbf{Videos}} \\ 
\midrule
UCF101~\cite{soomro2012ucf101} & 109.11 & - & - \\ 
DAVIS~\cite{perazzi2016benchmark} & 0.36 & - & - \\ 
HMDB~\cite{kuehne2011hmdb} & 55.41 & - & - \\ 
ActivityNet~\cite{caba2015activitynet} & 380.63 & - & - \\ 
Moments in Time~\cite{monfort2019moments} & 2979.00 & - & - \\ 
Multi-moments in Time~\cite{monfort2021multi} & 4124.04 & - & - \\ 
Co3D~\cite{reizenstein2021common} & 228.75 & - & - \\ 
Charades v1~\cite{sigurdsson2016hollywood} & 241.53 & - & - \\ 
Something-something v2~\cite{goyal2017something} & 904.57 & - & - \\ 
YouCook~\cite{das2013thousand} & 3.14 & - & - \\ 
Kinetics 700~\cite{carreira2019short} & 7092.04 & - & - \\ 
MSR-VTT~\cite{xu2016msr} & 57.34 & - & - \\ 
Youtube VOS~\cite{xu2018youtube} & 63.70 & - & - \\ 
jester~\cite{materzynska2019jester} & 606.47 & - & - \\ 
diving48~\cite{li2018resound} & 150.73 & - & - \\ 
MultiSports~\cite{li2021multisports} & 78.44 & - & - \\ 
CharadesEgo~\cite{sigurdsson2018charades} & 193.06 & - & - \\ 
AVA~\cite{murray2012ava} & 117.96 & - & - \\ 
Ego4D~\cite{grauman2022ego4d} & 1152.12 & - & - \\ 
\midrule
\multicolumn{4}{|c|}{\textbf{Videos with Annotations}} \\ 
\midrule
VIPSeg~\cite{miao2022large} & 64.47 & Video Panoptic Segmentation & Ground Truth \\ 
Hand14K~\cite{fathi2011learning} & 1.96 & Hand Segmentation  & Ground Truth \\ 
AVA~\cite{murray2012ava}    & 122.88 & Video Detection & Ground Truth \\ 
JHMDB~\cite{Jhuang:ICCV:2013}  & 19.00 & Optical Flow   & Ground Truth \\ 
JHMDB~\cite{Jhuang:ICCV:2013}  & 37.92 & Video Human Pose   & Ground Truth \\
\midrule
\multicolumn{4}{|c|}{\textbf{Synthetic 3D Views}} \\ 
\midrule
Objaverse~\cite{deitke2023objaverse} Rendered Multiviews & 217.85 & - & - \\ 
\bottomrule
\end{tabular}

\caption{Data sources of single images, images with annotations, videos and videos with annotations contained in UVDv1. In building the training data for \ours, we source annotations from a large number of datasets  covering a diverge set of vision tasks. In addition to the ground truth annotations, we also leverage model-generated annotation to further broaden our diversity.}
\label{tab:UVDv1}
\end{table*}

\subsection{Summary of Dataset Distribution in UVD}

The Unified Vision Dataset (UVD) encompasses a diverse array of visual data, aggregating over 430 billion tokens. The distribution of these tokens across various categories underscores the dataset's extensive coverage, see Figure~\ref{fig:token_distribution}:

\begin{figure}
   \centering
   \includegraphics[width=\linewidth]{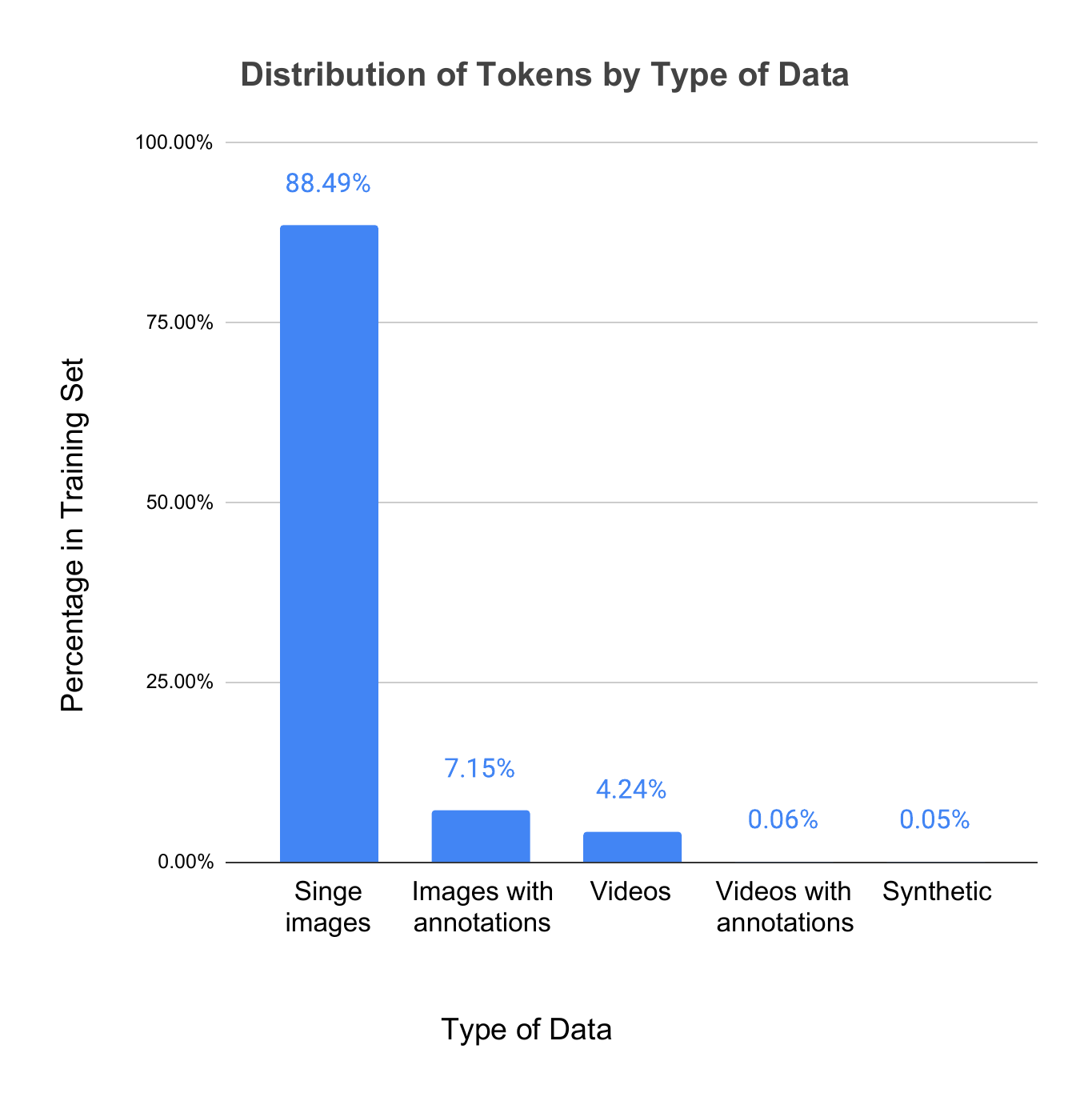}
   \caption{Tokens distribution of our training dataset. The majority of our training data comes from the single images of LAION, with the rest taking only 10\%.}
   \label{fig:token_distribution}
\end{figure}

    \paragraph{Single Images (88.49\%; 380.69 billion tokens)}: This category, featuring datasets like LAION~\cite{laion}, is the largest, providing a vast collection of unannotated images suitable for a wide range of applications, particularly in unsupervised learning.

    \paragraph{Images with Annotations (7.15\%; 30.78 billion tokens)}: Including prominent datasets such as ImageNet 1K~\cite{deng2009imagenet} and COCO~\cite{lin2014microsoft}, this segment offers annotated images for image classification, object detection, semantic segmentation etc.

    \paragraph{Videos (4.24\%; 18.26 billion tokens)}: Comprising datasets like UCF101~\cite{soomro2012ucf101} and Moments in Time~\cite{monfort2019moments}, this category provides unannotated video content, ideal for general video analysis and unsupervised learning in dynamic scenes.

    \paragraph{Videos with Annotations (0.06\%; 0.25 billion tokens)}: Though smaller in token count, this category is significant, with datasets like VIPSeg~\cite{miao2022large} and Hand14K~\cite{fathi2011learning} offering annotated videos for specific tasks like video segmentation and human pose estimation.

    \paragraph{Synthetic 3D Views (0.05\%; 0.22 billion tokens)}: Datasets like Objaverse~\cite{deitke2023objaverse} in this category cater to advanced 3D vision tasks, providing synthetic 3D views for cutting-edge research.

Overall, UVDv1's rich composition, with its extensive token array, positions it as a comprehensive resource for various tasks in computer vision, from basic image processing to complex analyses in video and 3D data.

\subsection{Details of Constructing Video Visual Sentences}

We implemented specific tokenization strategies for each video dataset, taking into account their unique characteristics and contents. These tailored tokenization processes, inclusive of epoch details, ensure a comprehensive and diverse representation of each dataset's unique video content.

    \paragraph{Something-something v2~\cite{goyal2017something}:} Tokenized with strides of 4 and 7, capturing sequences of 16 frames. Random starting points were used for each of the 10 epochs to ensure diversity in human-object interactions.

     \paragraph{CO3D~\cite{reizenstein2021common}:} Focused on 3D objects, tokenized with strides of 4 or 8 frames. Each sequence used 1 or 2 shots, with random starts in each epoch to capture object depth and detail.

    \paragraph{Ego4D~\cite{grauman2022ego4d}:} Strides of 12, 24, and 36 were employed, each sequence consisting of 16 frames. Randomization of starting points was implemented over 10 epochs to capture a range of egocentric activities.

     \paragraph{Charades v1~\cite{sigurdsson2016hollywood}:} Tokenized using strides of 10, 20, and 30 for 16-frame sequences. Random starting points across 2 epochs captured diverse narrative scenes.

    \paragraph{Kinetics 700~\cite{carreira2019short}:} Employed strides of 8 and 24, with each sequence capturing 16 frames. Random starts in each epoch over 10 epochs were used to represent a broad spectrum of human activities.

    \paragraph{Diving48~\cite{li2018resound}:} Strides of 2 and 4 for tokenization, capturing 32-frame sequences to detail diving techniques. Random starting points were utilized across all epochs for comprehensive motion analysis.

     \paragraph{AVA~\cite{murray2012ava}:} This dataset was tokenized with strides of 10 and 20, each sequence consisting of 16 frames. Random starts for sequences were used in each of the 50 epochs to capture varied human actions.

     \paragraph{Jester~\cite{materzynska2019jester}:} Tokenized to capture the subtlety of hand gestures with 16-frame sequences. Randomization in the starting points was employed to enhance gesture diversity.

    \paragraph{YouCook~\cite{das2013thousand}:} Tokenized with strides of 10, 20, and 30, each sequence comprising 16 frames. Random starting points over 4 epochs were used to capture a variety of cooking procedures.

    \paragraph{CharadesEgo~\cite{sigurdsson2018charades}:} Focused on first-person narratives, tokenized using strides of 10, 20, and 30 for 16-frame sequences over 2 epochs.

    \paragraph{YouTube VOS~\cite{xu2018youtube}:} Tokenized using strides of 2, 4, and 8, focusing on detailed object movements within 16-frame sequences over 2 epochs.

     \paragraph{MultiSports~\cite{li2021multisports}:} Captured sports actions with strides of 4, 8, and 12 for 16-frame sequences across 3 epochs.

     \paragraph{ActivityNet~\cite{caba2015activitynet}:} Tokenized with strides of 5, 10, and 15, capturing 16 frames per sequence over 4 epochs to represent a wide range of activities.

    \paragraph{Hand14K~\cite{fathi2011learning}:} Focused on hand gesture recognition, tokenized with sequences of 16 frames, capturing detailed hand movements over multiple epochs.

    \paragraph{Moments in Time~\cite{monfort2019moments}:} Captured a wide array of activities and phenomena with a stride of 0, considering the short length of the videos, over multiple epochs.

    \paragraph{Multi-Moments in Time~\cite{monfort2021multi}:} An extension of Moments in Time, tokenized with strides of 0, 2, and 4 for different runs, each sequence comprising 16 frames to capture simultaneous actions over multiple epochs.
    

\section{Additional Results}
\subsection{Sequential Prompting}

\begin{figure*}
    \centering
    \includegraphics[width=0.95\linewidth]{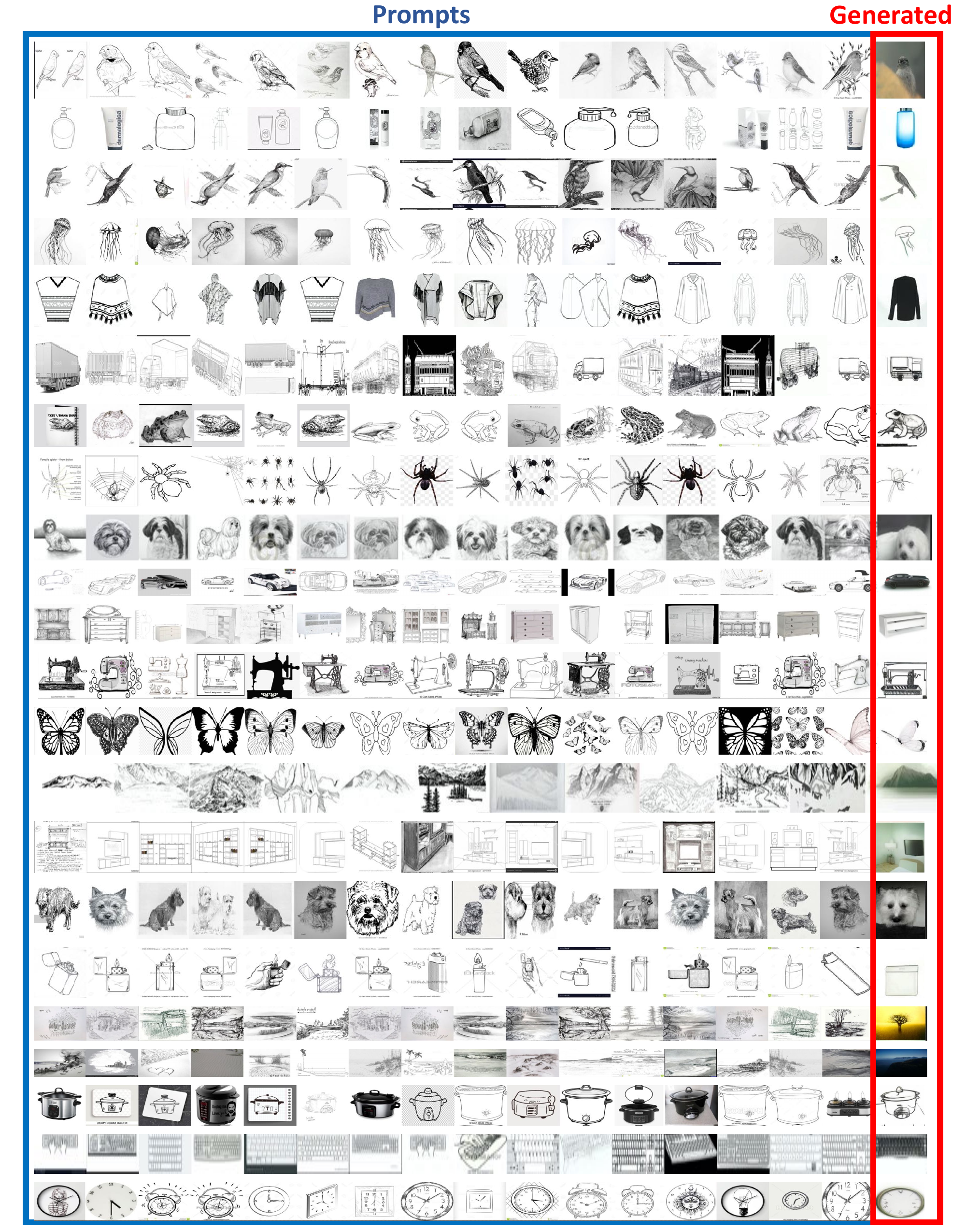}
    \caption{\textbf{Sketch understanding.} We construct visual sentences by sequences of sketches. LVM is asked to predict the next image.}
    \label{fig:sketch}
\end{figure*}

Additional results for sequential prompting are presented, including:

\paragraph{Sketch Understanding:} Figure \ref{fig:sketch} illustrates the model's capability in interpreting hand-drawn sketches from ImageNet-Sketch~\cite{wang2019learning}. We construct visual sentence from a sequence of 15 images from ImageNet-Sketch~\cite{wang2019learning} and then ask the model to predict the subsequent image. This method evaluates LVM's proficiency in interpreting and understanding hand-drawn sketches.

\begin{figure*}
    \centering
    \includegraphics[width=0.95\linewidth]{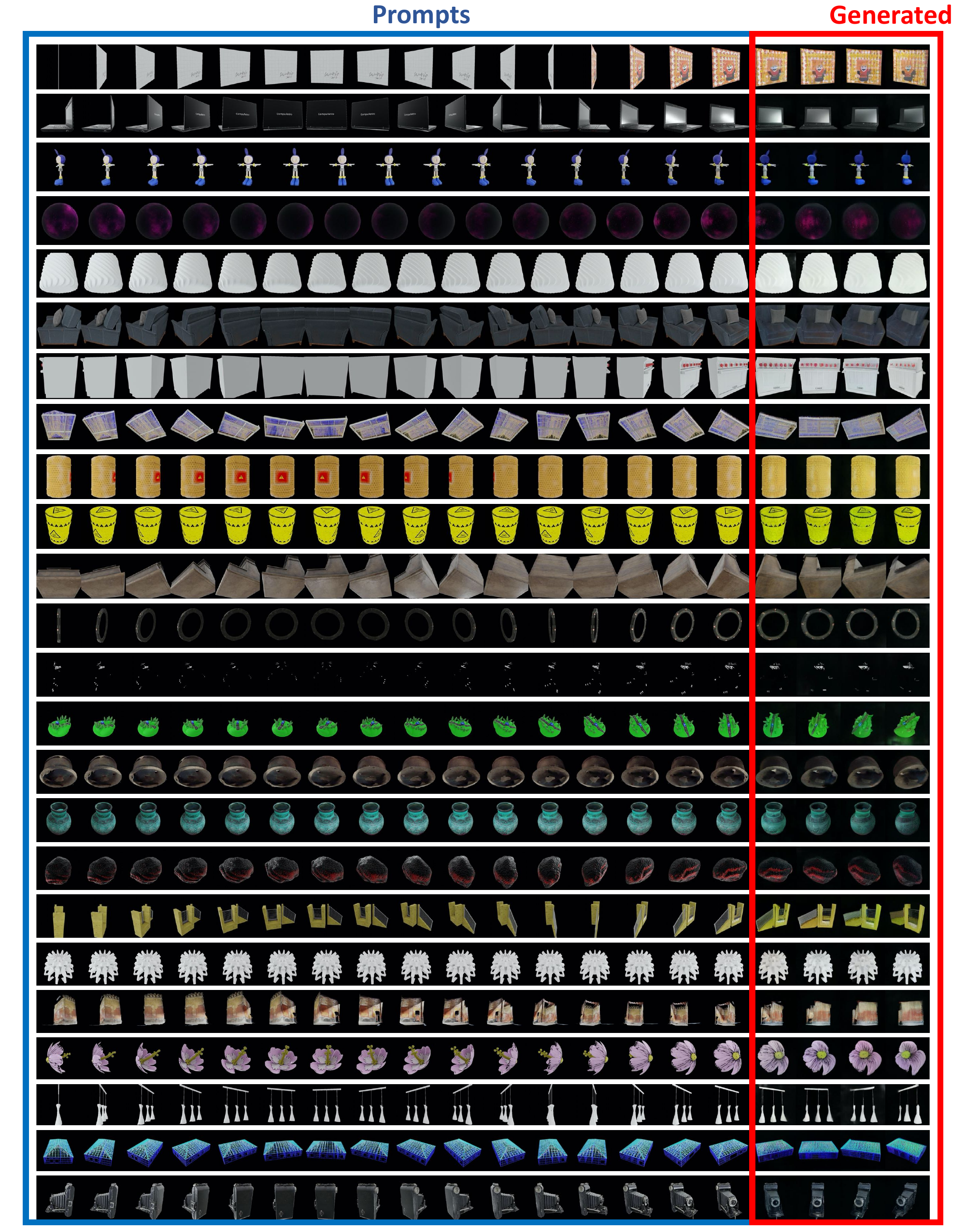}
    \caption{\textbf{3D Rotation about arbitrary axes.} We construct visual sentences by rotating images. LVM is asked to predict the next 4 views.}
    \label{fig:3d}
\end{figure*}

\paragraph{3D Rotation about arbitrary axes:} In our evaluation set for Objaverse, we adopt a range of unseen objects to test LVM's ability to handle arbitrary axis rotation. The model predicts the next 4 images based on a visual sentence of 16 images.  As illustrated in Figure~\ref{fig:3d}, LVM demonstrates its capacity to reason about the direction of spatial rotation based on the context provided by the prompt, leading to reasonable predictions. For this tasks, LVM exhibits 11.8 as in perplexity.

\begin{figure*}
    \centering
    \includegraphics[width=\linewidth]{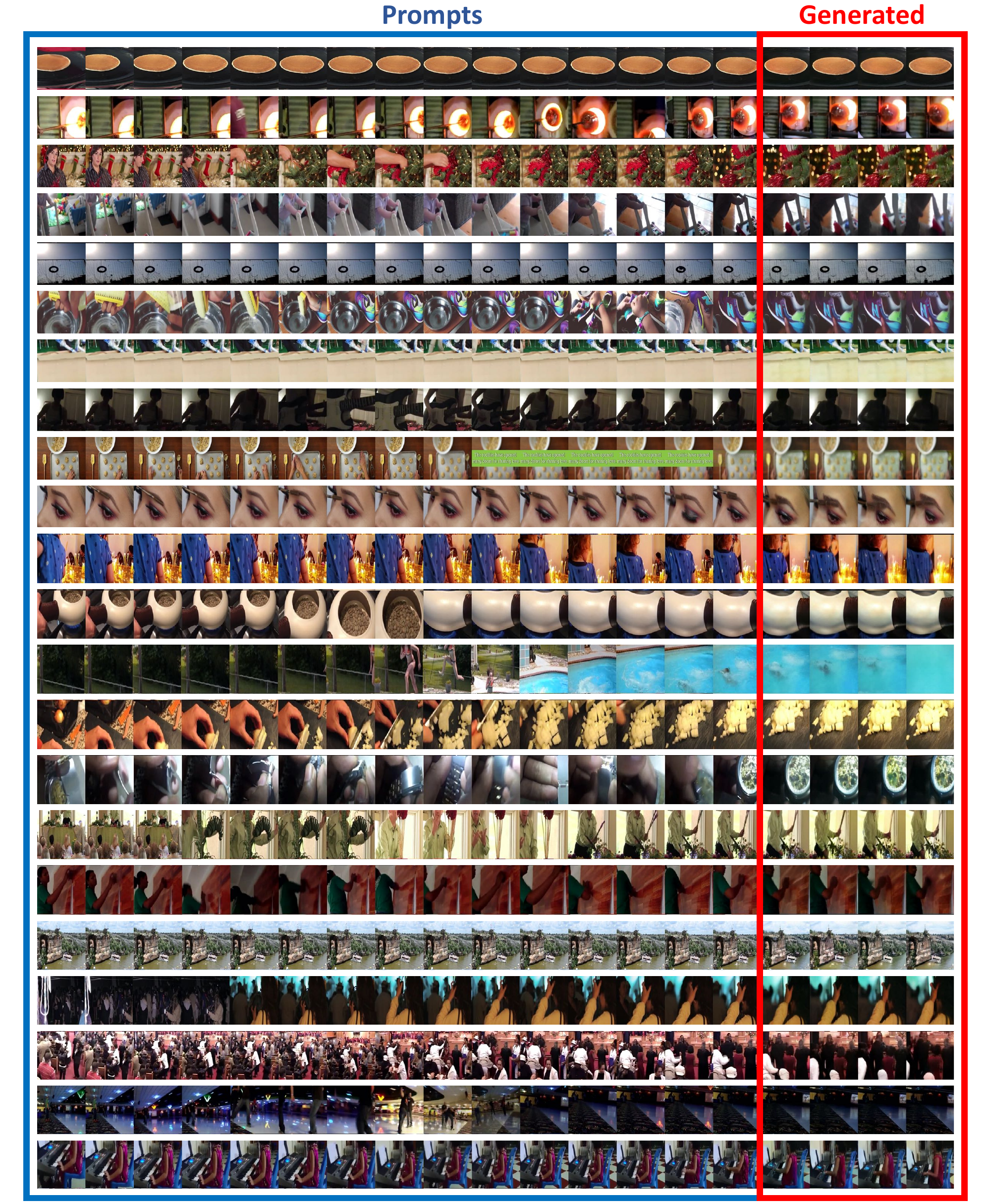}
    \caption{\textbf{Frames prediction.} We construct the visual sentence by sequences of frames. LVM is asked to predict the next 4 frames.}
    \label{fig:k700_1}
\end{figure*}

\begin{figure*}
    \centering
    \includegraphics[width=\linewidth]{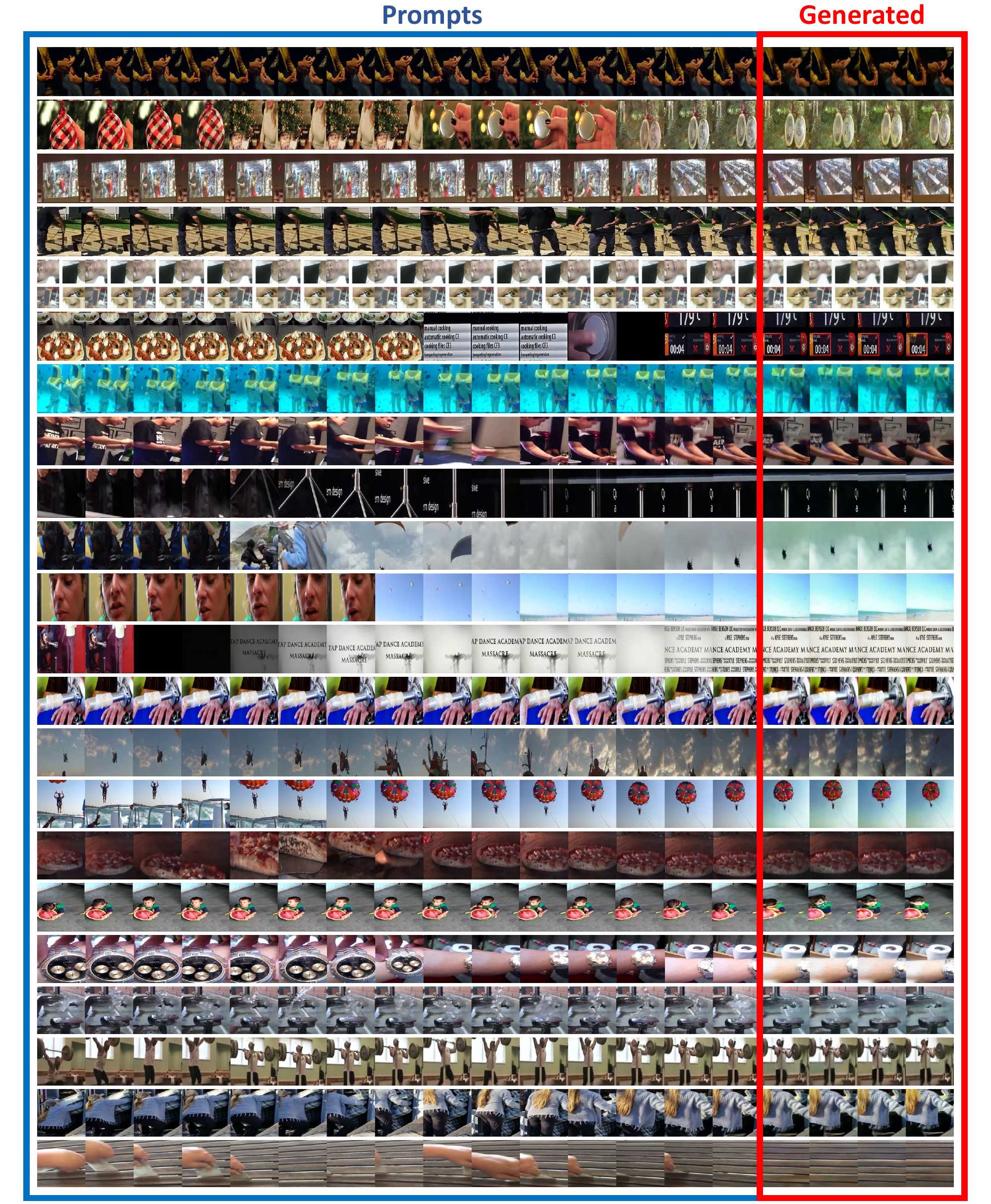}
    \caption{\textbf{Frames prediction.} We construct the visual sentence by sequences of frames. LVM is asked to predict the next 4 frames.}
    \label{fig:k700_2}
\end{figure*}

\begin{figure*}
    \centering
    \includegraphics[width=\linewidth]{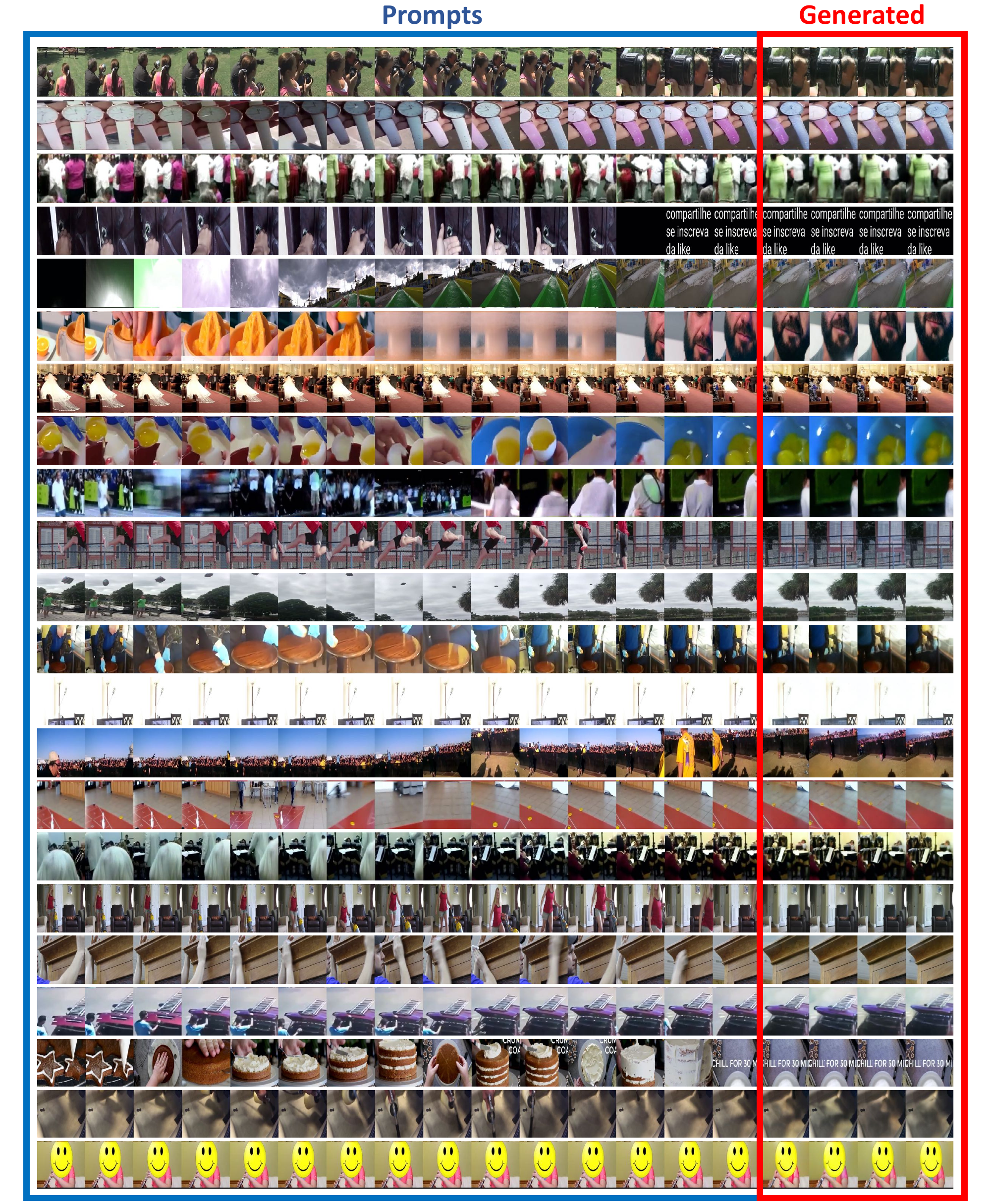}
    \caption{\textbf{Frames prediction.}  We construct the visual sentence by sequences of frames. LVM is asked to predict the next 4 frames.}
    \label{fig:k700_3}
\end{figure*}

\begin{figure*}
    \centering
    \includegraphics[width=\linewidth]{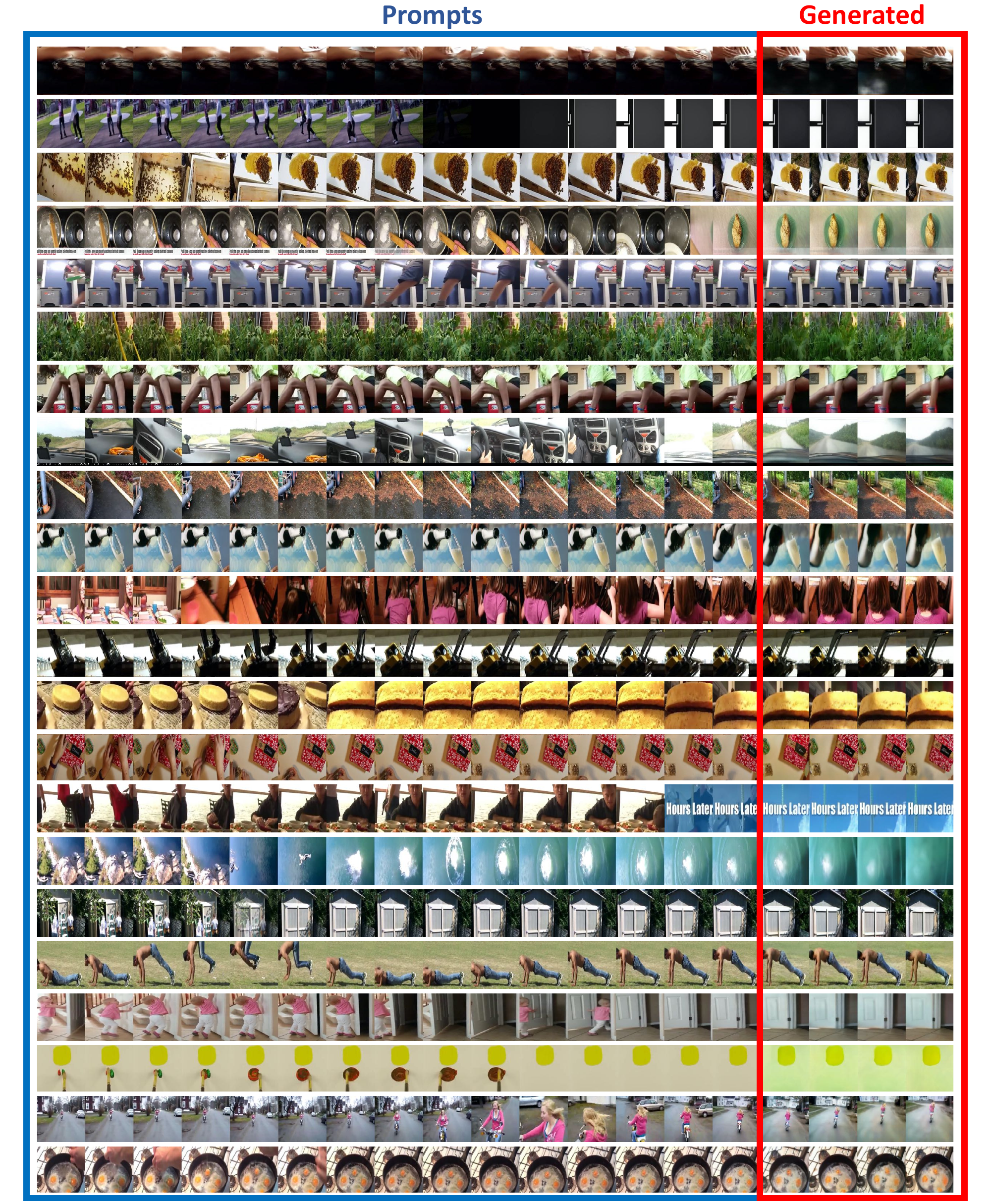}
    \caption{\textbf{Frames prediction.}  We construct the visual sentence by sequences of frames. LVM is asked to predict the next 4 frames.}
    \label{fig:k700_4}
\end{figure*}

\begin{figure*}
    \centering
    \includegraphics[width=\linewidth]{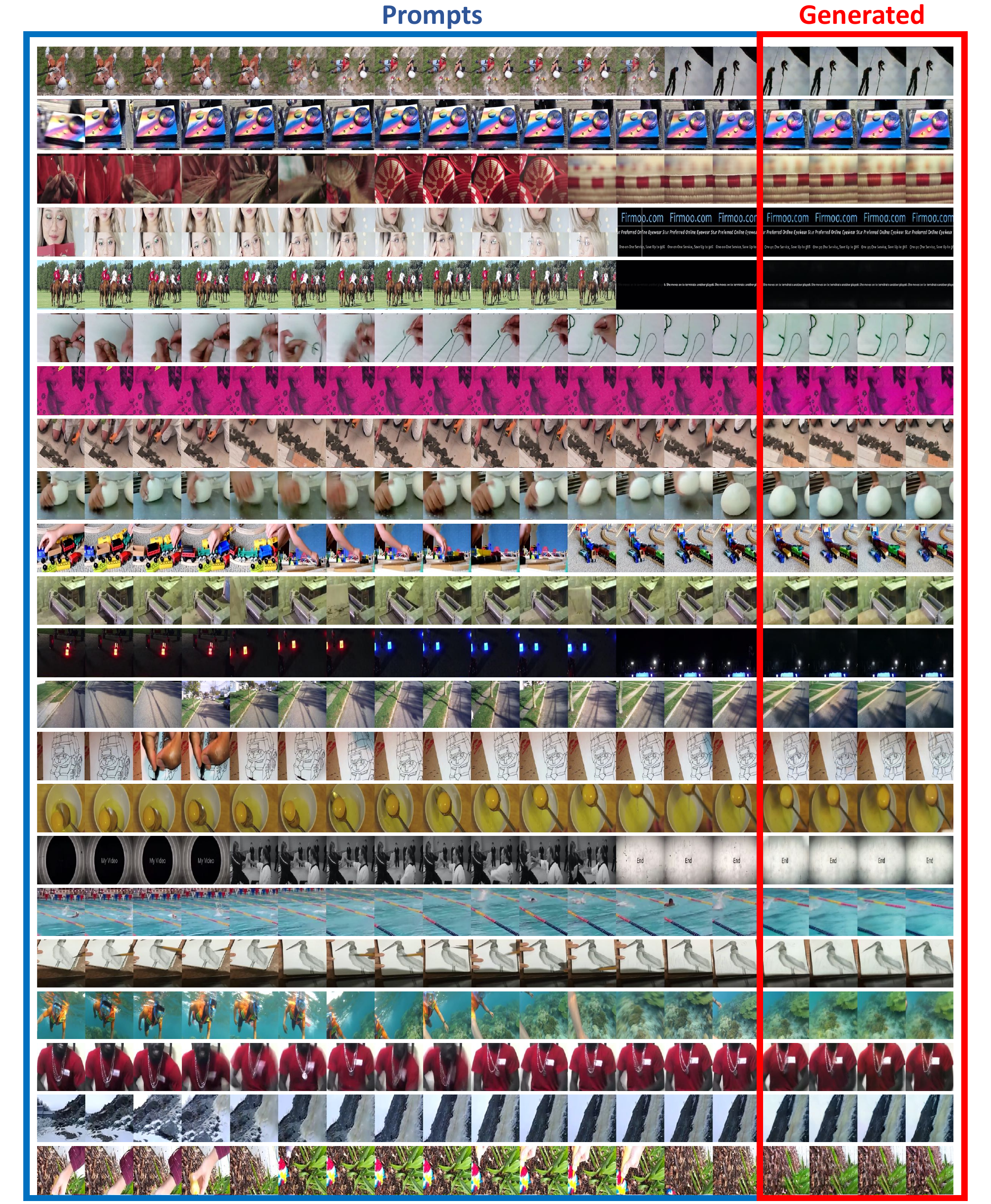}
    \caption{\textbf{Frames prediction.}  We construct the visual sentence by sequences of frames. LVM is asked to predict the next 4 frames.}
    \label{fig:k700_5}
\end{figure*}

\begin{figure*}
    \centering
    \includegraphics[width=\linewidth]{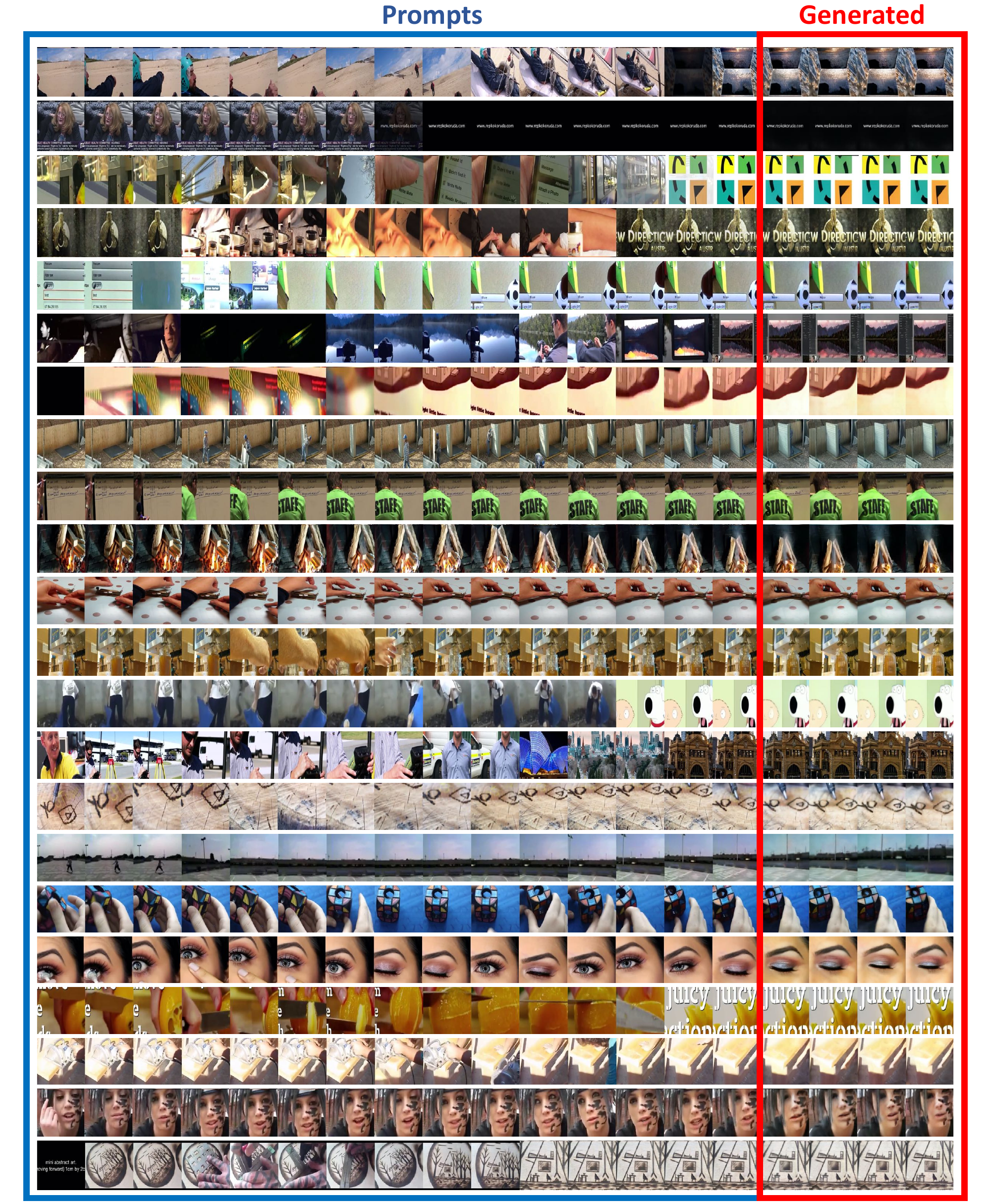}
    \caption{\textbf{Frames prediction.}  We construct the visual sentence by sequences of frames. LVM is asked to predict the next 4 frames.}
    \label{fig:k700_6}
\end{figure*}

\paragraph{Frames Prediction:} Figures \ref{fig:k700_1} to \ref{fig:k700_6} demonstrate frame prediction using the evaluation set from Kinetics 700 dataset. The model predicts the next 4 frames based on a visual sentence of 16 frames. The Fréchet Inception Distance (FID) score for single-frame prediction conditioned on 15 frames is 21.018, indicating the LVM's proficiency in understanding spatial and temporal dynamics.

\subsection{Analogy Prompting}
Further results for analogy prompting in various contexts are provided, highlighting the model's adaptability and understanding in different scenarios.

\begin{figure*}
    \centering
    \includegraphics[width=0.95\linewidth]{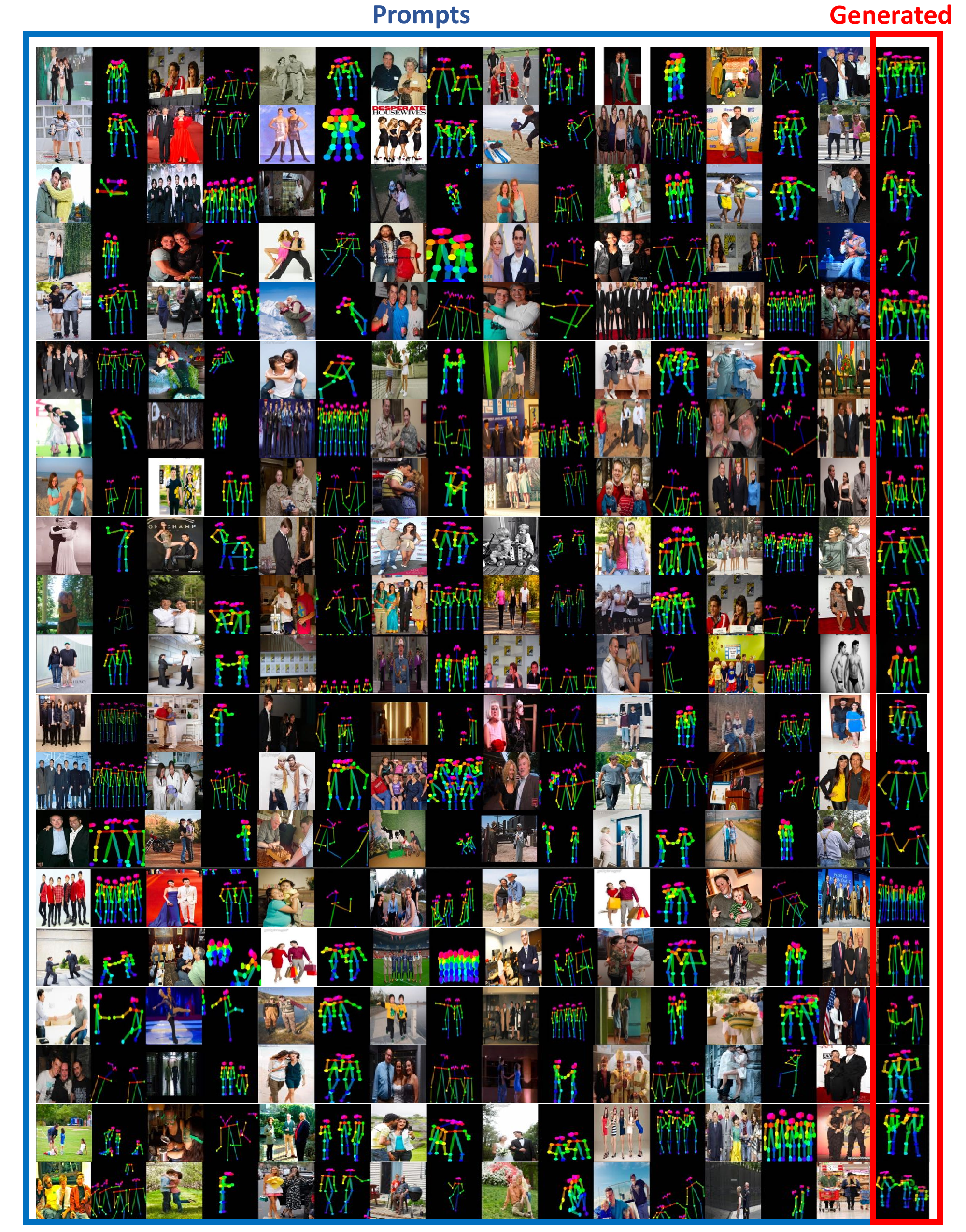}
    \caption{\textbf{Human keypoint detection.} We construct the visual sentence by ``image-to-joints'' analogy prompting from LVMHP~\cite{li2017multiple} dataset. LVM is asked to predict the skeleton of all humans in the image. }
    \label{fig:kp_1}
\end{figure*}

\paragraph{Pose Estimation Analogy:}
In Figure \ref{fig:kp_1}, the pose estimation analogy is constructed using the visual sentence of ``image-to-joint'', where the model predicts poses from given images. This assesses the model's ability to interpret analogy pairs and understand human poses and joint relationships.

\begin{figure*}
    \centering
    \includegraphics[width=0.95\linewidth]{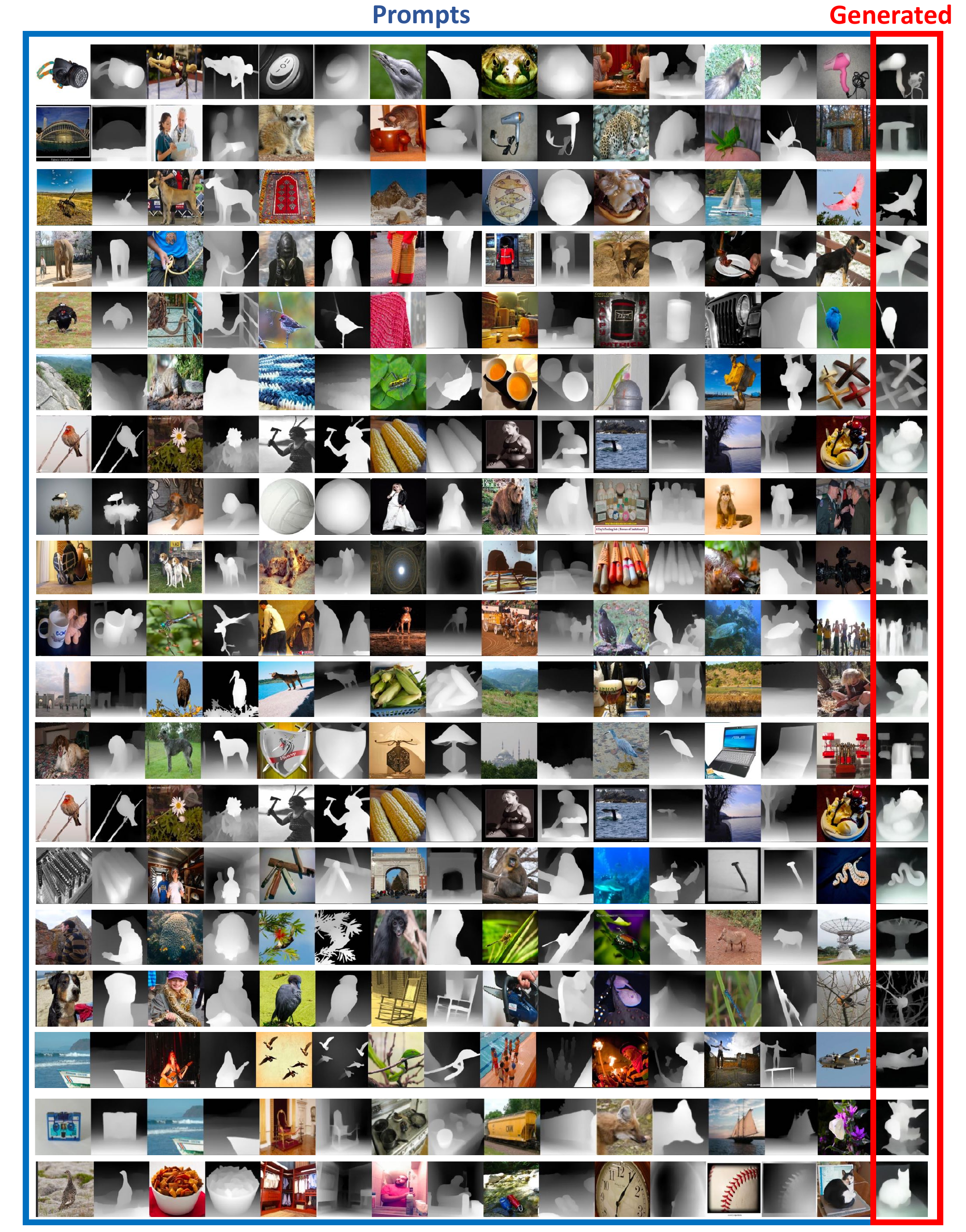}
    \caption{\textbf{Depth Estimation.} We construct the visual sentence by ``image-to-depth image'' analogy prompting from ImageNet validation set. LVM is asked to predict the depth map.}
    \label{fig:depth_1}
\end{figure*}

\paragraph{Depth Estimation Analogy:}
Figure \ref{fig:depth_1} presents the ``image-to-depth'' analogy for depth estimation. The visualizations utilize the validation set from \cite{deng2009imagenet}, whose annotations are generated by DPT  \cite{ranftl2021vision}, and re-normalised to [-1, 1] following~\cite{prismer}.

\begin{figure*}
    \centering
    \includegraphics[width=0.95\linewidth]{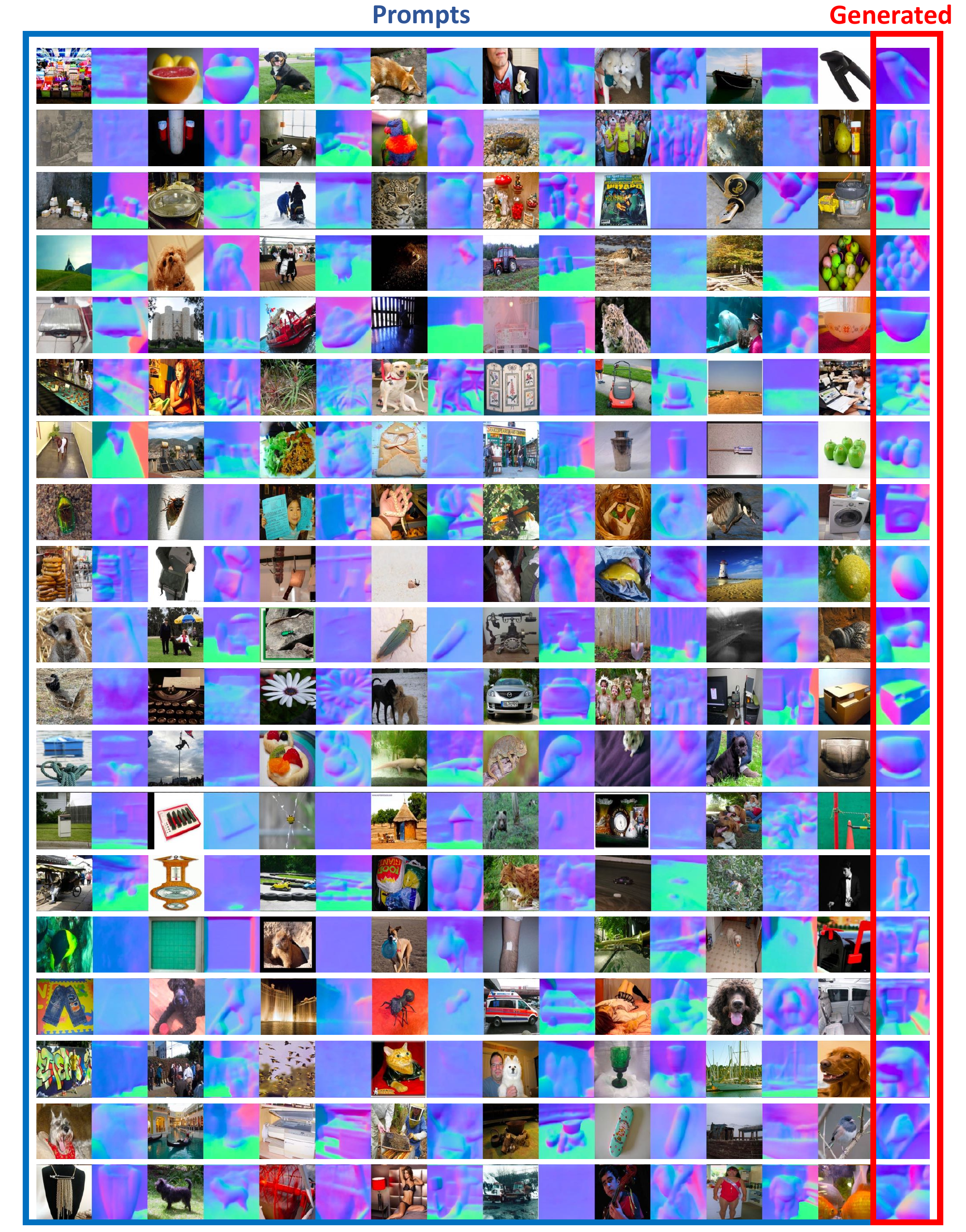}
    \caption{\textbf{Surface Normal Estimation.} We construct the visual sentence by ``image-to-surface normal image'' analogy prompting from ImageNet validation set. LVM is asked to predict the surface normal estimation map. }
    \label{fig:normal_1}
\end{figure*}

\paragraph{Surface Normal Estimation Analogy:}
The ``image-to-surface normal image'' analogy is depicted in Figure \ref{fig:normal_1}. This analogy tests the model's depth of understanding of 3D structures from 2D data. Despite inaccuracies in some normal surface images from the prompts, our model shows notable robustness and generalization.

\begin{figure*}
    \centering
    \includegraphics[width=0.95\linewidth]{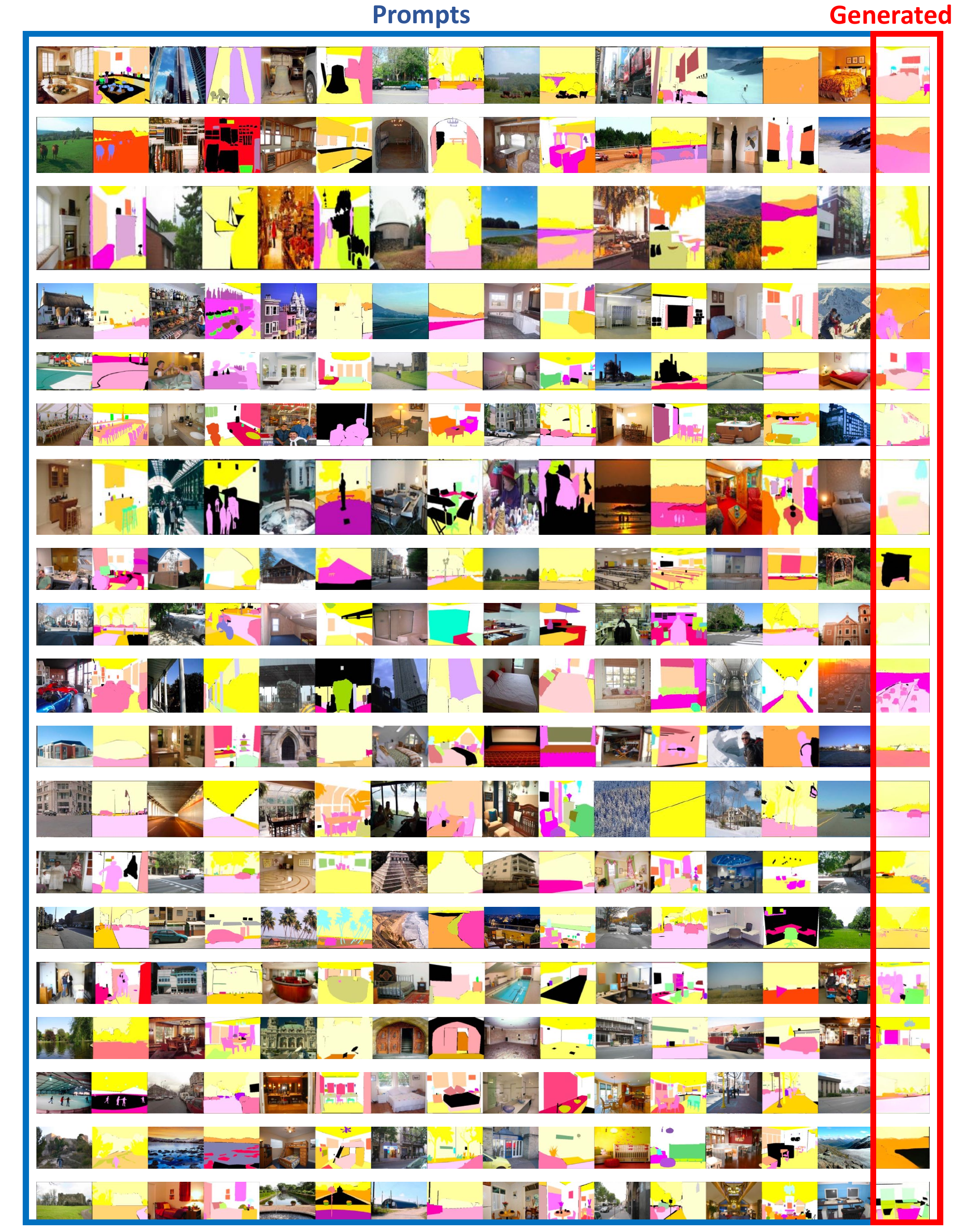}
    \caption{\textbf{Semantic Segmentation.} We construct the visual sentence by ``image-to-segmentation'' analogy prompting from ADE 20K validation set.  LVM is asked to predict semantic segmentation color map. }
    \label{fig:seg_1}
\end{figure*}

\paragraph{Semantic Segmentation Analogy:} Results for the ``image-to-segmentation'' analogy are shown in Figure \ref{fig:seg_1}, emphasizing semantic segmentation. The visualizations are based on the validation set from ADE20K~\cite{zhou2019semantic}.

\begin{figure*}
    \centering
    \includegraphics[width=0.95\linewidth]{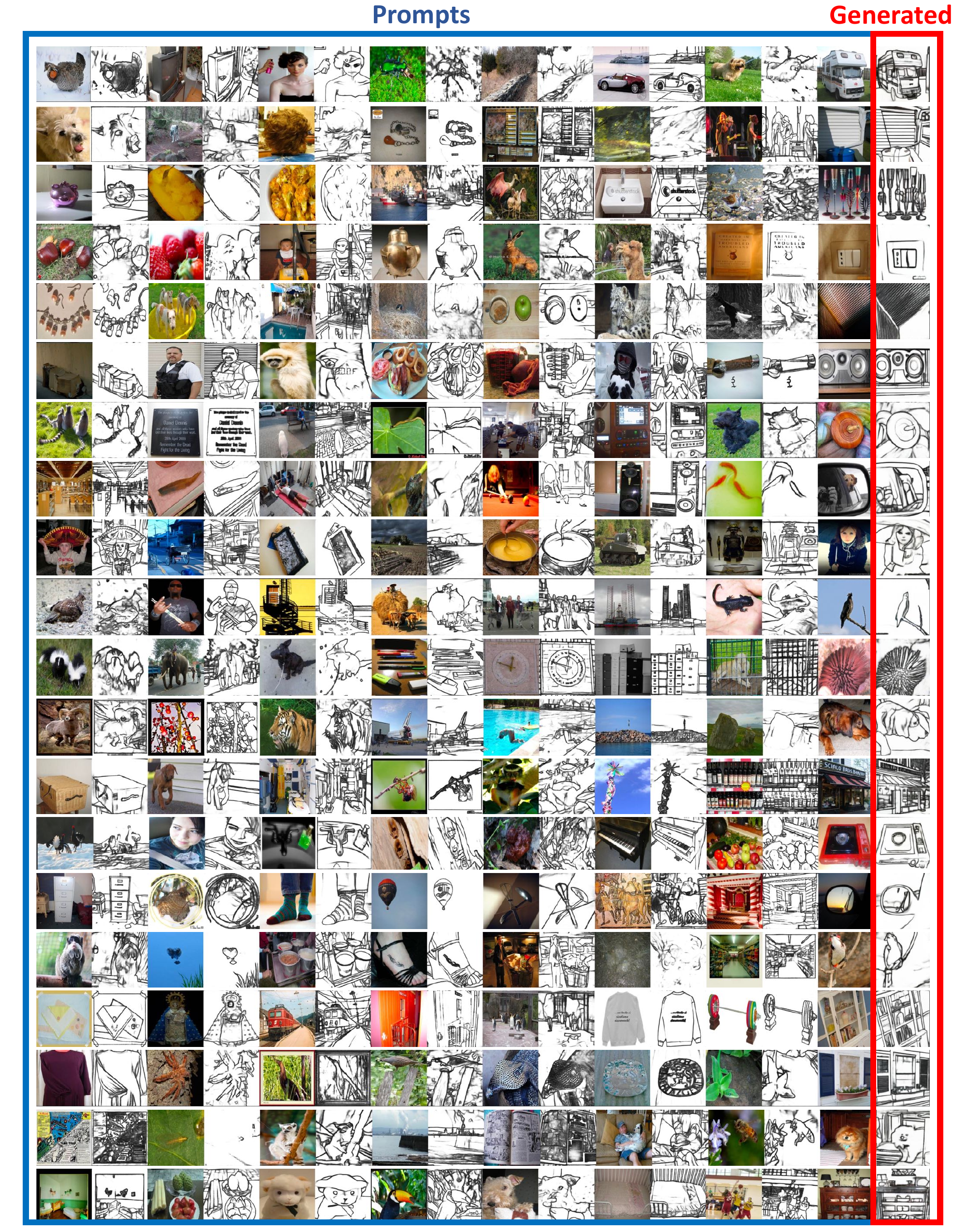}
    \caption{\textbf{Edge Detection.}  We construct the visual sentence by ``image-to-edge'' analogy prompting from ImageNet validation set. LVM is asked to predict the edge map given a new image.}
    \label{fig:edge_1}
\end{figure*}

\paragraph{Edge Detection Analogy:}
Results for the ``image-to-edge'' analogy are shown in Figure \ref{fig:edge_1}, emphasizing edge detection. The visualizations are based on the validation set from \cite{deng2009imagenet}, annotated using DexiNed~\cite{xsoria2020dexined}.

\begin{figure*}
    \centering
    \includegraphics[width=0.95\linewidth]{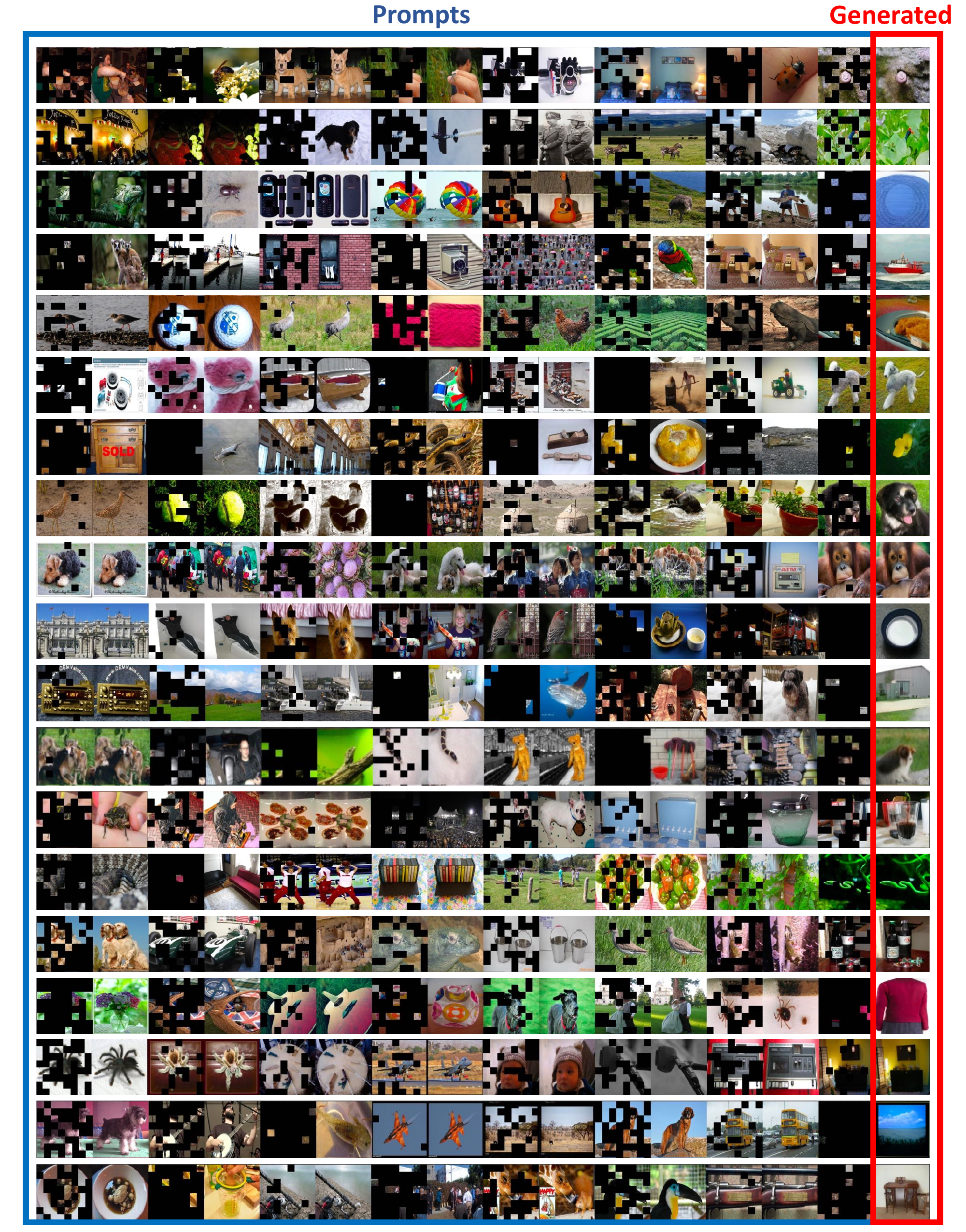}
    \caption{\textbf{Inpainting.}  We construct the visual sentence by ``partially masked image-to-image'' analogy prompting from ImageNet validation set. LVM is asked to reconstruct the pixel of the masked region given a new partially masked image. }
    \label{fig:inpainting_1}
\end{figure*}

\paragraph{Image Inpainting Analogy:}
In Figure \ref{fig:inpainting_1}, the ``partially masked image-to-image'' analogy is explored, demonstrating the model's capabilities in image inpainting. The model is challenged with different mask ratios, showing significant semantic understanding, as evidenced by a Mean Squared Error (MSE) of 0.106.

\begin{figure*}
    \centering
    \includegraphics[width=0.95\linewidth]{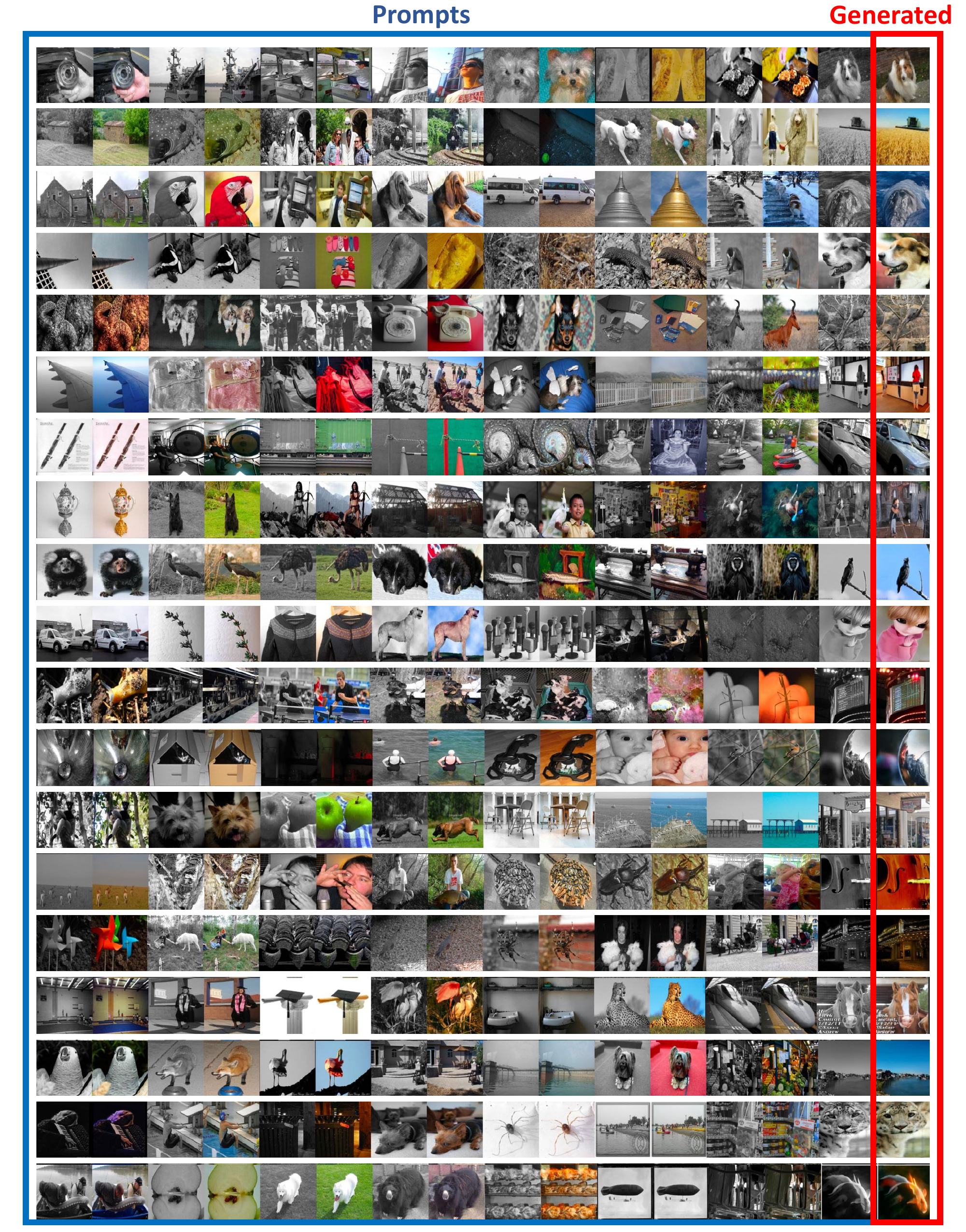}
    \caption{\textbf{Colorization.} We construct the visual sentence by ``gray-scale image-to-image'' analogy prompting from ImageNet validation set. LVM is asked to colorize the image given a new gray-scale image.}
    \label{fig:colorization_1}
\end{figure*}

\paragraph{Image Colorization Analogy:}
Figure \ref{fig:colorization_1} shows the ``gray-scale image-to-image'' analogy for image colorization. This test showcases the model's ability to handle complex image scenarios, with an MSE of 0.51.

\begin{figure*}
    \centering
    \includegraphics[width=0.95\linewidth]{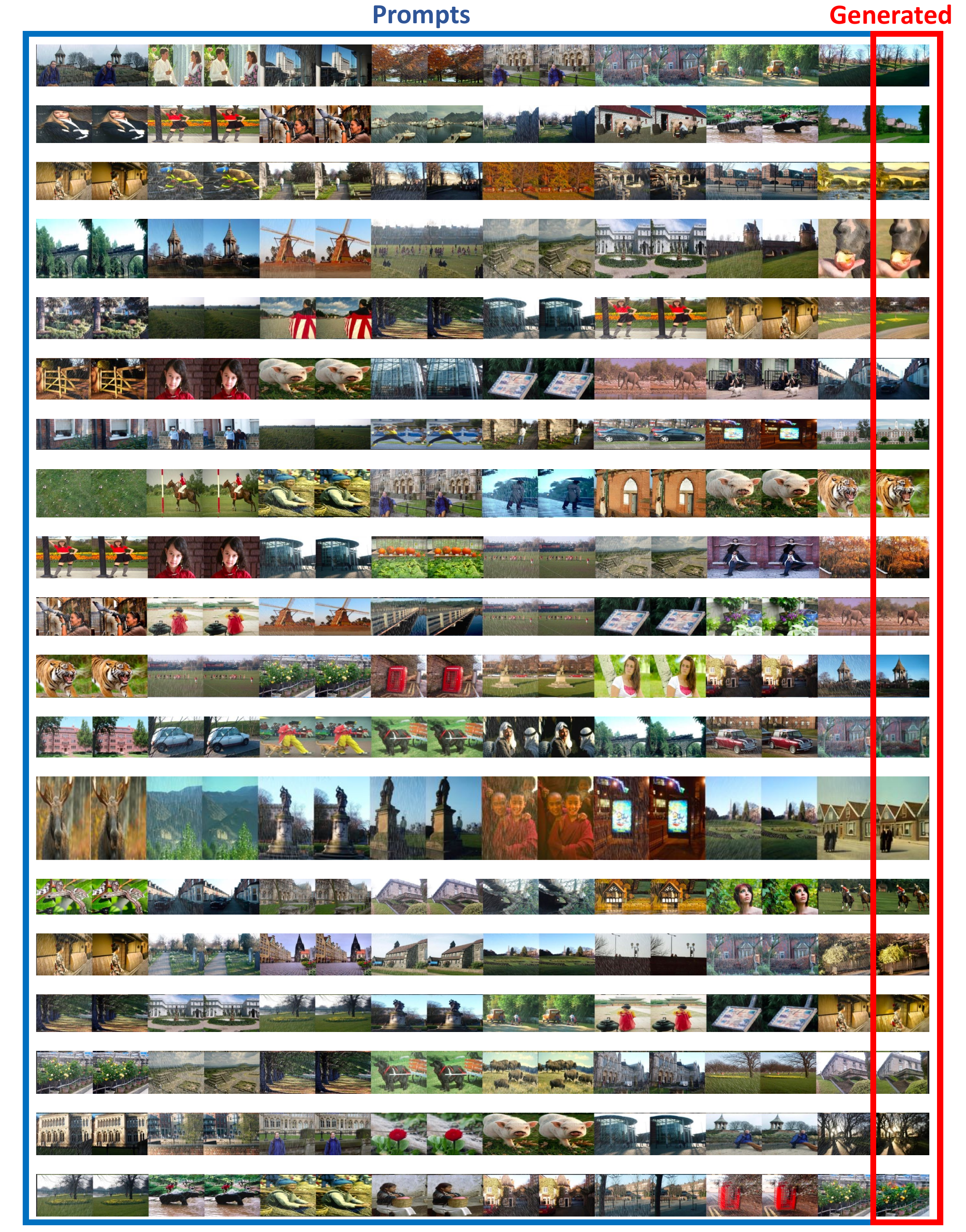}
    \caption{\textbf{Derain.} We construct the visual sentence by ``rainy image-to-image'' analogy prompting from DID-MDN~\cite{derain_zhang_2018} validation set. LVM is asked to derain the image.}
    \label{fig:derain}
\end{figure*}

\paragraph{Derain Analogy:}
Figure \ref{fig:derain} shows the ``rainy image-to-image'' analogy for image deraining. 


\clearpage



\end{document}